\documentclass{article} 
\usepackage[preprint]{colm2026_conference}

\usepackage{microtype}
\usepackage{hyperref}
\usepackage{url}
\usepackage{booktabs}

\usepackage{amsmath}
\usepackage{multirow}
\usepackage{graphicx}

\usepackage{algorithm}
\usepackage{algpseudocode}
\usepackage{tcolorbox}

\usepackage{lineno}

\definecolor{darkblue}{rgb}{0, 0, 0.5}
\hypersetup{colorlinks=true, citecolor=darkblue, linkcolor=darkblue, urlcolor=darkblue}

\title{From Execution to Education: A Bloom-Aligned Framework for Measuring Educational Control in LLMs}

\author{Yi Zhang \& Julia Rayz \\
School of Applied Creative Computing\\
Purdue University\\
West Lafayette, IN 47907, USA \\
\texttt{\{zhan3050,jtaylor1\}@purdue.edu} \\
}

\begin{document}

\ifcolmsubmission
\linenumbers
\fi

\maketitle

\begin{abstract}
We introduce a Bloom-aligned framework for measuring educational control in Large Language Models (LLMs): the ability to preserve a task's instructional intent while shifting its cognitive demand toward specified learning objectives. We apply this framework to programming tasks in computer science education to study the gap between solving tasks and adapting them for learners. Using revised Bloom's Taxonomy as an operational scale of cognitive demand, we evaluate two intervention settings: general difficulty control, where models are asked to make tasks harder or easier, and Bloom's control, where models are asked to target higher or lower Bloom's levels. We evaluate a matched Qwen3-Next model pair, comparing Qwen3-Next-80B-A3B-Instruct with Qwen3-Coder-Next across 2,520 tasks from three benchmarks. The framework reveals a robust directional asymmetry: both models reliably increase cognitive demand, but struggle to lower it. We further characterize these outcomes with semantic-delta clustering and layer-wise Fisher's Discriminant Ratio probing. Within this controlled comparison, the general model shows clearer middle-layer separability for both general difficulty and Bloom-control contrasts, whereas the coder model shows weaker separability for general difficulty and a deeper peak for Bloom-control contrasts. These results show that strong execution performance does not automatically entail Bloom-aligned educational control.
\end{abstract}

\section{Introduction}
\label{sec:intro}

Recent advancements in Large Language Models (LLMs) have demonstrated significant success in extending their capabilities to complex software engineering tasks. As these models become deeply integrated into real-world applications, they are increasingly deployed beyond code completion to act as collaborative educational tools \citep{houEffectsGenerativeAI2024} and intelligent tutors \citep{hkudsDeepTutorAIpoweredPersonalized2025, yuMOOCMAICReimagine2026}. However, excelling at functional execution benchmarks illustrates that a model can be a proficient problem solver without being an effective educator. The Zone of Proximal Development (ZPD) \citep{vygotskyMindSocietyDevelopment1980} provides useful background for this distinction: education requires dynamically adjusting difficulty so that tasks remain challenging enough to encourage learning without becoming unproductive. We therefore study \emph{educational control}: the task-level ability to preserve instructional intent while shifting cognitive demand toward specified learning objectives. In this paper, we operationalize educational control with revised Bloom's Taxonomy \citep{andersonTaxonomyLearningTeaching2001}, which provides an ordered structure of cognitive processes. Programming tasks provide a controlled testbed for observing the gap between execution performance and educational adaptation.

To measure LLM proficiency in software engineering, the research community has developed a vast and diverse ecosystem of benchmarks, ranging from static, function-level evaluations like HumanEval \citep{chenEvaluatingLargeLanguage2021} to complex, repository-level frameworks such as SWE-bench \citep{jimenezSWEbenchCanLanguage2024}. While these execution metrics are essential for ensuring that generated code is valid, they are pedagogically shallow because they treat code generation as a pass-or-fail outcome and ignore the underlying cognitive load required to solve the problem. When LLMs are applied directly to computer science education, optimizing solely for immediate, functionally correct solutions frequently causes an ``illusion of learning'' \citep{pratherWideningGapBenefits2024}. To leverage these models as educational tools, we must look beyond standard execution benchmarks and implement extra considerations to ensure they guide students through the productive struggle necessary to build independent problem-solving skills.

While prior work establishes that certain LLMs are able to evaluate the cognitive complexity of existing tasks \citep{huberLLMsMeetBloom`s2025}, a critical gap remains in understanding how they actively respond to educational controls during generation. Specifically, there is limited evidence about whether models can preserve a task's instructional intent while moving the required cognitive operation along a learning taxonomy, rather than only changing surface form. It remains unclear whether specialized training on software engineering data alters a model's capacity to handle general difficulty control and Bloom-targeted educational control, or if these behaviors reflect broader training-distribution biases. To investigate task control behavior and its representational correlates, we analyze Qwen3-Next-80B-A3B-Instruct \citep{yangQwen3TechnicalReport2025} as the general model and Qwen3-Coder-Next \citep{caoQwen3CoderNextTechnicalReport2026} as the coder model. We evaluate how these models handle difficulty prompts requesting harder or easier tasks, measure their responses to Bloom-targeted prompts targeting higher or lower educational levels, and characterize the resulting patterns by mapping external linguistic shifts and probing the separability of internal neural representations.

This paper makes three primary contributions. (1) We introduce a Bloom-aligned framework for measuring educational control in LLMs, built around matched difficulty interventions, Bloom-targeted interventions, and two metrics: Observed Cognitive Shift (OCS) and Target Zone Accuracy (TZA). (2) We apply this framework to a controlled general/coder model pair across 2,520 programming tasks, showing a consistent directional asymmetry: the models reliably raise cognitive demand but often fail to lower it. (3) We connect these behavioral outcomes to external lexical mutation strategies and layer-wise representation diagnostics, revealing distinct mutation patterns and separability profiles across the matched models.

\section{Related Work}
\label{sec:related_work}

This work sits at the intersection of three strands of prior research: execution-centric evaluation of LLMs on programming tasks, pedagogical frameworks for cognitive control, and representation diagnostics for internal model states.

\subsection{LLMs in Computing Education}

The rapid adoption of LLMs has shifted software development paradigms and pushed the educational community to ask whether these systems can support learning rather than only synthesis. Much of the early evaluation literature still centers on execution correctness, with benchmarks such as HumanEval \citep{chenEvaluatingLargeLanguage2021} and MBPP \citep{austinProgramSynthesisLarge2021} treating code generation as a pass-or-fail problem. More recent benchmarks broaden the setting to contamination-aware or repository-level tasks, including LiveCodeBench \citep{jainLiveCodeBenchHolisticContamination2024} and SWE-bench \citep{jimenezSWEbenchCanLanguage2024}, but they remain execution-centric in what they measure. The pedagogical concern is that a correct solution can still bypass productive struggle; prior work in computing education shows that direct answers can create an illusion of learning even when the final code is valid \citep{kazemitabaarExploringDesignSpace2025, pratherWideningGapBenefits2024}. Controllable educational generation moves beyond static evaluation and asks whether models can intentionally shape the cognitive level of what they produce. BloomWise \citep{zoumpoulidiBloomWiseEnhancingProblemSolving2024} uses Bloom-inspired prompting to guide LLM reasoning through cognitive stages, while chain-of-hints systems study how models can scaffold learners toward a solution through static or adaptive hints \citep{jangraDesigningEvaluatingChainofHints2026}.

\subsection{Educational Frameworks and LLM Evaluation}

Researchers use various models to measure learning and reasoning, such as outcome-focused structures \citep{biggsEvaluatingQualityLearning1982, zhangLLMBasedFrameworkSimulating2025} and depth of knowledge frameworks \citep{webbCriteriaAlignmentExpectations1997, yuRecallReasoningAutomated2025}. Revised Bloom's Taxonomy \citep{andersonTaxonomyLearningTeaching2001} organizes learning objectives into six ordered cognitive processes: Remember, Understand, Apply, Analyze, Evaluate, and Create. Recent studies apply Bloom's to find specific AI limits. For example, current research highlights LLM failures in high-level evaluation tasks \citep{huberLLMsMeetBloom`s2025} and reveals distinct performance drops in automated program repair \citep{maBloomAPRBloomsTaxonomybased2025}.

\subsection{Representation Diagnostics}
Mechanistic interpretability and representation analysis have developed a substantial toolkit for analyzing how Transformers represent and transform information. The residual stream \citep{elhage2021mathematical} and logit lens \citep{nostalgebraistInterpretingGPTLogit2020} showed that intermediate states can often be decoded long before the final layer. The linear representation hypothesis \citep{parkLinearRepresentationHypothesis2024}, together with linear probing \citep{alainUnderstandingIntermediateLayers2018, belinkovProbingClassifiersPromises2022}, treats concepts as directions in activation space, and later work showed that these directions often shift from broad abstractions to more task-specific features with depth \citep{gurneeLanguageModelsRepresent2024, guptaHowLLMsUse2026, skeanLayerLayerUncovering2025}. The field also developed a complementary interventional literature. The causal gap \citep{ravichanderProbingProbingParadigm2021} motivated amnesic probing \citep{elazarAmnesicProbingBehavioral2021}, activation patching \citep{mengLocatingEditingFactual2022}, and circuit tracing \citep{wangInterpretabilityWildCircuit2023}; circuit-based work then clarified induction heads \citep{olssonIncontextLearningInduction2022}, sparse feature circuits \citep{marksSparseFeatureCircuits2025}, and code-relevant routing patterns \citep{liLlmsBuildWorld2024}. In parallel, sparse autoencoders \citep{brickenMonosemanticityDecomposingLanguage2023} and representation engineering \citep{zouRepresentationEngineeringTopDown2025} turned these ideas into controllable directions, including code- and education-focused analyses \citep{yinNeuronGuidedInterpretationCode2025, raimondiMechanisticInterpretabilityCognitive2026}.

\section{Methodology}
\label{sec:method}

Our methodology follows a three-stage pipeline. We first generate task interventions under two instruction settings: general difficulty control and Bloom's control. The Bloom-control setting has target zones, while the difficulty setting is used observationally to compare how harder/easier requests behave on the Bloom scale. We then evaluate whether those interventions actually move tasks toward the intended cognitive level for Bloom-targeted prompts, or how they shift along the Bloom scale for difficulty prompts. Finally, we examine both the lexical changes and the internal representations associated with those shifts. The exact prompt templates and decoding settings are documented in Appendix~\ref{app:settings}.

\subsection{Model Architectures and Task Datasets}
We evaluate two open-weights models from the Qwen3-Next family: Qwen3-Next-80B-A3B-Instruct \citep{yangQwen3TechnicalReport2025} as the general model and Qwen3-Coder-Next \citep{caoQwen3CoderNextTechnicalReport2026} as the coder model. Both models use the same 48-layer architecture and tokenizer, which gives a controlled comparison between a general-purpose instruction model and a code-specialized model while reducing architectural and tokenization confounds. We chose this pair because, at the time of study, it was the closest open-weights general/coder pair we could identify with matched architecture and tokenizer.

To cover a range of coding behaviors, we use 2,520 programming tasks sampled from three benchmarks. BigCodeBench \citep{zhuoBigCodeBenchBenchmarkingCode2025} contains 1,140 Python function call tasks, LiveCodeBench v5 \citep{jainLiveCodeBenchHolisticContamination2024} contains 880 algorithmic problem solving tasks, and SWE-Bench-Verified \citep{openaiIntroducingSWEbenchVerified2024} contains 500 repository-level software engineering tasks. For each task, we generate four independent interventions under two instruction settings: general difficulty control, where the task is made ``Harder'' or ``Easier'', and Bloom's control, where the task is targeted at a ``Higher'' (Evaluate/Create) or ``Lower'' (Remember/Understand) Bloom's Taxonomy level. All four interventions use the same template and decoding settings so comparisons are not driven by direction-specific prompt tuning. The Harder/Easier setting provides an observational comparison showing how general difficulty requests behave on the Bloom scale alongside Bloom control.

\subsection{Behavioral Evaluation of Cognitive Mutations}
We employ an LLM-as-a-judge framework using \texttt{claude-3.5-haiku} \citep{anthropicClaude35Haiku2024} to evaluate the success of these interventions. This model was selected based on recent work validating its performance on Bloom's Taxonomy classification for coding tasks \citep{Zhang_Rayz_2026}. Specifically, it achieved an agreement rate (Gwet's AC2 of 0.95) against human consensus on a 150-question validation subset drawn from the same three benchmark sources, providing a reliable and scalable method to categorize cognitive level. Because generated mutations may differ from the original validation distribution, we use the judge primarily to compare relative movement across matched intervention conditions rather than to assert that every absolute label is definitive. We quantify the outcome of each intervention with two customized metrics. The exact prompt templates for both judge and interventions are documented in Appendix~\ref{app:settings}. Observed Cognitive Shift (OCS) measures the average movement along Bloom's Taxonomy, treating the taxonomy as an ordinal scale from 1 (Remember) to 6 (Create). It is formally defined in Equation \ref{eq:ocs}:
\begin{equation}
    \label{eq:ocs}
    \text{OCS} = \frac{1}{N} \sum_{i=1}^{N} \left( L_{\text{mut}}^{(i)} - L_{\text{orig}}^{(i)} \right)
\end{equation}
where $L_{\text{orig}}^{(i)}$ is the original cognitive level and $L_{\text{mut}}^{(i)}$ is the mutated level for task $i$.

Target Zone Accuracy (TZA) complements OCS by measuring precision. It checks whether a model lands within the Bloom's level zone, as defined in Equation \ref{eq:tza}:
\begin{equation}
    \label{eq:tza}
    \text{TZA} = \frac{1}{N} \sum_{i=1}^{N} \mathbb{1}\left(L_{\text{mut}}^{(i)} \in Z^{(i)}\right)
\end{equation}
where $\mathbb{1}$ is an indicator function, and $Z^{(i)}$ are the targeted Bloom's levels we specify. 

\subsection{External Linguistic Strategy Analysis}
To understand how models change cognitive complexity, we examine the textual strategies they use during generation. The mutated task still contains most of the original programming content, so we cannot interpret it in isolation. Instead, we adapt the BERTopic pipeline \citep{grootendorstBERTopicNeuralTopic2022} with a semantic delta calculation that isolates the mutation from the shared task context.

First, we encode the original and mutated tasks in a dense semantic space using the \texttt{all-mpnet-base-v2} embedding model \citep{reimersSentenceBERTSentenceEmbeddings2019}. Following the Linear Representation Hypothesis \citep{parkLinearRepresentationHypothesis2024}, we model the semantic intent of each mutation as a directional vector in the embedding space (Equation \ref{eq:diff}):
\begin{equation}
    \label{eq:diff}
    \mathbf{E}_{\text{diff}} = \mathbf{E}_{\text{mutated}} - \mathbf{E}_{\text{original}}
\end{equation}
This semantic delta removes the shared background context of the coding problem. The resulting $\mathbf{E}_{\text{diff}}$ vector then represents only the linguistic shift introduced by the model. Once the delta vectors are isolated, we identify consistent behavioral patterns across the generated tasks. We project the high-dimensional differences into a lower-dimensional manifold with UMAP \citep{mcinnesUMAPUniformManifold2020} and then cluster the result with HDBSCAN \citep{campelloDensityBasedClusteringBased2013}. This density-based clustering is useful because it finds natural groupings of variable density while filtering out noisy or irregular modifications as outliers.

After grouping similar mutation strategies, we extract the vocabulary that defines them. Standard topic-modeling pipelines typically use class-based TF-IDF (c-TF-IDF), but that approach is better suited to comparing clusters against one another than to comparing each mutation against its original input. Our goal is the latter: we want the lexical injection relative to the starting state. We therefore use the Weighted Log Odds Ratio with an informative Dirichlet prior \citep{monroeFightinWordsLexical2008}. For every cluster, this metric contrasts the vocabulary of the mutated tasks with that of their original counterparts. The full pipeline is summarized in Algorithm \ref{alg:clustering}. Because this analysis is exploratory, we interpret the clusters as recurring linguistic strategies rather than as exhaustive or causal explanations of the mutations.

\begin{algorithm}[ht!]
\caption{Semantic Delta Clustering for Keywords Extraction}
\label{alg:clustering}
\begin{algorithmic}[1]
\Require $\text{original\_tasks}, \text{mutated\_tasks}$
\Ensure $\text{mutation\_clusters}, \text{distinctive\_keywords}$
\For{each task pair $(\text{orig}, \text{mut})$}
    \State $\mathbf{E}_{\text{orig}} \gets \text{SBERT\_all\_mpnet\_base\_v2}(\text{orig})$
    \State $\mathbf{E}_{\text{mut}} \gets \text{SBERT\_all\_mpnet\_base\_v2}(\text{mut})$
    \State $\mathbf{E}_{\text{diff}} \gets \mathbf{E}_{\text{mut}} - \mathbf{E}_{\text{orig}}$
\EndFor
\State $\text{Reduced\_Dims} \gets \text{UMAP}(\mathbf{E}_{\text{diff}}, \text{n\_components}=5)$
\State $\text{Clusters} \gets \text{HDBSCAN}(\text{Reduced\_Dims}, \text{min\_cluster\_size}=15)$
\For{each cluster $C$ in $\text{Clusters}$}
    \State $C_{\text{mut}} \gets \text{mutated tasks within } C$
    \State $C_{\text{orig}} \gets \text{original tasks within } C$
    \State $\text{Keywords}[C] \gets \text{WeightedLogOddsRatio}(C_{\text{mut}}, C_{\text{orig}})$
\EndFor
\State \Return $\text{Clusters}, \text{Keywords}$
\end{algorithmic}
\end{algorithm}

\subsection{Internal Representation Diagnostic Probing}
To examine the internal representations associated with these linguistic shifts, we probe the models' activations across all network layers. Capturing the activations at the last instruction token allows us to measure where the instruction contrast becomes most linearly separable across depth.

We use Fisher's Discriminant Ratio (FDR) instead of a traditional classifier. FDR is a parameter-free measure of linear separability that compares between-class variance to within-class variance, as shown in Equation \ref{eq:fdr}:
\begin{equation}
    \label{eq:fdr}
    \text{FDR} = \frac{(\mu_1 - \mu_2)^2}{\sigma_1^2 + \sigma_2^2}
\end{equation}
where $\mu$ and $\sigma^2$ represent the mean and variance of the activations for each intervention setting.

By computing FDR across all layers, we identify the depth where the instruction classes are most separable under this linear diagnostic. To visualize this separation, we compute a mean-difference direction ($\mathbf{V}_{\text{diff}} = \mu_{\text{higher}} - \mu_{\text{lower}}$) at the peak layer and project the activations onto that axis, as detailed in Algorithm \ref{alg:probing}.

\begin{algorithm}[ht!]
\caption{Internal Representation Probing}
\label{alg:probing}
\begin{algorithmic}[1]
\Require $\text{generator\_model}, \text{dataset}, \text{labels}$
\Ensure $\text{max\_fdr\_layer}, \text{latent\_projections}$
\For{each layer $L$ in $\text{generator\_model}$}
    \State $\text{Act}[L] \gets \text{CaptureResidualStream}(\text{dataset}, \text{token}=\text{LAST\_INST})$
    \State $\text{FDR}[L] \gets \frac{(\mu(\text{Act}_{\text{pos}}) - \mu(\text{Act}_{\text{neg}}))^2}{\sigma^2(\text{Act}_{\text{pos}}) + \sigma^2(\text{Act}_{\text{neg}})}$
\EndFor
\State $\text{Best\_Layer} \gets \text{argmax}(\text{FDR})$
\State $\mathbf{V}_{\text{diff}} \gets \mu(\text{Act}[\text{Best\_Layer}]_{\text{pos}}) - \mu(\text{Act}[\text{Best\_Layer}]_{\text{neg}})$
\State $\text{Unit\_Vector} \gets \frac{\mathbf{V}_{\text{diff}}}{\|\mathbf{V}_{\text{diff}}\|}$
\State $\text{Projections} \gets \text{Act}[\text{Best\_Layer}] \cdot \text{Unit\_Vector}$
\State \Return $\text{Best\_Layer}, \text{Projections}$
\end{algorithmic}
\end{algorithm}

\section{Behavioral Results of Task Mutation}
\label{sec:behavioral_results}

We first report observed directional shifts in Bloom's level under both difficulty and Bloom-targeted prompts, then measure target adherence under the targeted Bloom's levels setting.

\subsection{Observed Cognitive Shift}
To explore how models interpret general difficulty compared to specific Bloom's targets, we calculate the Observed Cognitive Shift (OCS) under both prompting strategies. Table \ref{tab:ocs_combined} summarizes these directional shifts across benchmarks.

\begin{table}[ht!]
\centering
\resizebox{0.8\textwidth}{!}{%
\begin{tabular}{@{}llcccc@{}}
\toprule
\multirow{2}{*}{Model} & \multirow{2}{*}{Benchmark} & \multicolumn{2}{c}{Difficulty} & \multicolumn{2}{c}{Bloom's Level} \\ \cmidrule(l){3-6} 
 &  & Harder & Easier & Higher$\uparrow$ & Lower$\downarrow$ \\ \midrule
\multirow{4}{*}{Coder Model} & Overall & 1.582 & 0.715 & 1.495 & 0.103 \\
 & \quad BigCodeBench & 2.097 & 1.120 & 1.910 & 0.332 \\
 & \quad SWE-Bench-Verified & 1.930 & 0.222 & 1.166 & -0.940 \\
 & \quad LiveCodeBench & 0.717 & 0.472 & 1.144 & 0.399 \\ \midrule
\multirow{4}{*}{General Model} & Overall & 1.762 & 0.194 & 2.146 & -0.134 \\
 & \quad BigCodeBench & 2.375 & 0.255 & 2.916 & 0.032 \\
 & \quad SWE-Bench-Verified & 2.226 & -0.362 & 1.810 & -1.250 \\
 & \quad LiveCodeBench & 0.705 & 0.432 & 1.341 & 0.285 \\ \bottomrule
\end{tabular}%
}
\caption{Observed Cognitive Shift (OCS). Arrows ($\uparrow$/$\downarrow$) indicate higher/lower is better for targeted Bloom's levels. Difficulty mutations are observational and have no defined target direction.}
\label{tab:ocs_combined}
\end{table}

When instructed to make a task ``Harder,'' both models demonstrate a strong upward cognitive shift (an average OCS of 1.582 for the coder model and 1.762 for the general model). These shifts mirror the behavior observed when models are explicitly prompted to target a ``Higher'' Bloom's level. This alignment suggests that, in this setting, general difficulty-increase requests tend to move tasks into higher cognitive tiers.

However, an asymmetry appears when attempting to reduce cognitive demand. Under the general ``Easier'' prompt, both models still produce positive OCS scores (0.715 and 0.194), showing that a request for lower surface difficulty does not necessarily yield lower Bloom-level demand. Comparing this to the targeted ``Lower'' Bloom's setting reveals that Bloom context slightly mitigates this issue (the general model achieved a marginal negative shift -0.134), but the overarching resistance remains. OCS scores reveal that the models more easily add cognitively demanding operations than remove or replace them with lower-level operations, under both general difficulty prompts and targeted Bloom's levels.

\subsection{Target Zone Accuracy}
We use Target Zone Accuracy (TZA) to measure exact adherence to the targeted Bloom's levels (Remember or Understand for ``Lower'', Evaluate or Create for ``Higher'').

\begin{table}[ht!]
\centering
\resizebox{0.7\textwidth}{!}{%
\begin{tabular}{@{}lllcc@{}}
\toprule
Model & Benchmark & N & Higher$\uparrow$ & Lower$\uparrow$ \\ \midrule
\multirow{4}{*}{Coder Model} & Overall & 2520 & 0.634 & 0.251 \\
 & \quad BigCodeBench & 1140 & 0.742 & 0.252 \\
 & \quad SWE-Bench-Verified & 500 & 0.766 & 0.630 \\
 & \quad LiveCodeBench & 880 & 0.419 & 0.035 \\ \midrule
\multirow{4}{*}{General Model} & Overall & 2520 & 0.792 & 0.292 \\
 & \quad BigCodeBench & 1140 & 0.979 & 0.246 \\
 & \quad SWE-Bench-Verified & 500 & 0.978 & 0.786 \\
 & \quad LiveCodeBench & 880 & 0.445 & 0.072 \\ \bottomrule
\end{tabular}%
}
\caption{Target Zone Accuracy (TZA) by model and benchmark. The upward arrows ($\uparrow$) indicate that a higher score is better.}
\label{tab:tza_combined}
\end{table}

As shown in Table \ref{tab:tza_combined}, both architectures perform well when targeting ``Higher'' levels. The general model succeeds 79.2\% of the time, outperforming the coder model's 63.4\% success rate. By contrast, targeting ``Lower'' levels exposes a clear weakness, with overall accuracy dropping below 30\% for both models.

The benchmark breakdown is not uniform. SWE-Bench-Verified stands out as the clearest case where simplification is comparatively easier, especially for the general model, while LiveCodeBench remains the most resistant setting for downward shifts. Rather than treating this as an exception that weakens the main result, we interpret it as evidence that repository-level tasks expose more room for local simplification than isolated algorithmic puzzles.

A closer look reveals a meaningful architectural divergence. Under the ``Higher'' command, the general model consistently moves tasks to the highest \textit{Create} tier, abstracting existing logic into new algorithmic design requirements. The coder model is more incremental: it often stops at the \textit{Evaluate} tier and tends to add constraints to existing code rather than asking for original creation.

\section{Linguistic and Representation-Diagnostic Results}
\label{sec:diagnostic_results}

To contextualize the behavioral outcomes, we examine the external lexical patterns used during generation alongside diagnostic probes of internal representations.

\subsection{External Linguistic Strategies}
We use semantic delta clustering to extract keywords introduced during task modification. Under difficulty intervention, the two models deploy noticeably different lexical strategies to manifest their upward bias. The general model forms nine distinct clusters in the ``Harder'' setting, employing varied terms such as \textit{optional}, \textit{bool}, and \textit{nested} to increase structural complexity in multiple ways. Conversely, the coder model concentrates its ``Harder'' mutations into just three tight clusters, focusing on input rules and error-handling constraints via words like \textit{valid}, \textit{handle}, \textit{compute}, and \textit{cases}.

This strategic divergence persists under the ``Higher'' intervention setting. The general model adapts its vocabulary to the specific benchmark domain. For algorithmic tasks in LiveCodeBench, it injects performance-oriented words such as \textit{algorithm}, \textit{optimal}, and \textit{complexity}. However, for repository-level challenges in SWE-Bench-Verified, it pivots to architectural terminology like \textit{design}, \textit{refactor}, and \textit{strategy}. This linguistic flexibility indicates an ability to reframe problems into entirely new solution spaces, aligning with the \textit{Create} tier of Bloom's Taxonomy.

The coder model demonstrates a more rigid lexical approach across all benchmarks. When adapting isolated algorithms or repository-level tasks, it repeatedly relies on inspection-focused terms like \textit{implementation}, \textit{evaluate}, \textit{behavior}, and \textit{uses}. Instead of requiring students to generate novel algorithms, it conservatively prompts them to analyze existing code execution. Comprehensive keyword tables mapping these strategies across all four intervention settings and three benchmarks are provided in Appendix~\ref{app:keywords} Tables~\ref{tab:keyword_general_he} through \ref{tab:keyword_coder_hl}.

The keyword analysis also helps explain the failures on ``Easier'' and ``Lower''. Neither model consistently deploys a vocabulary indicative of Bloom-level simplification. When attempting to lower cognitive load, they often fall back on structural or descriptive boilerplate, using words such as \textit{function}, \textit{takes}, \textit{write}, and \textit{list} for ``Easier'', and \textit{given}, \textit{following}, and \textit{does} for ``Lower''. This pattern suggests that the models may satisfy surface-form instructions by making tasks more explicit or more formatted while leaving the core reasoning operation intact. Appendix~\ref{app:downward_examples} provides concrete downward-mutation examples that separate surface formatting changes from actual Bloom-level changes.

\subsection{Internal Latent Representations}
\begin{figure}[ht]
\centering
\begin{minipage}{0.45\textwidth}
    \centering
    \includegraphics[width=\textwidth]{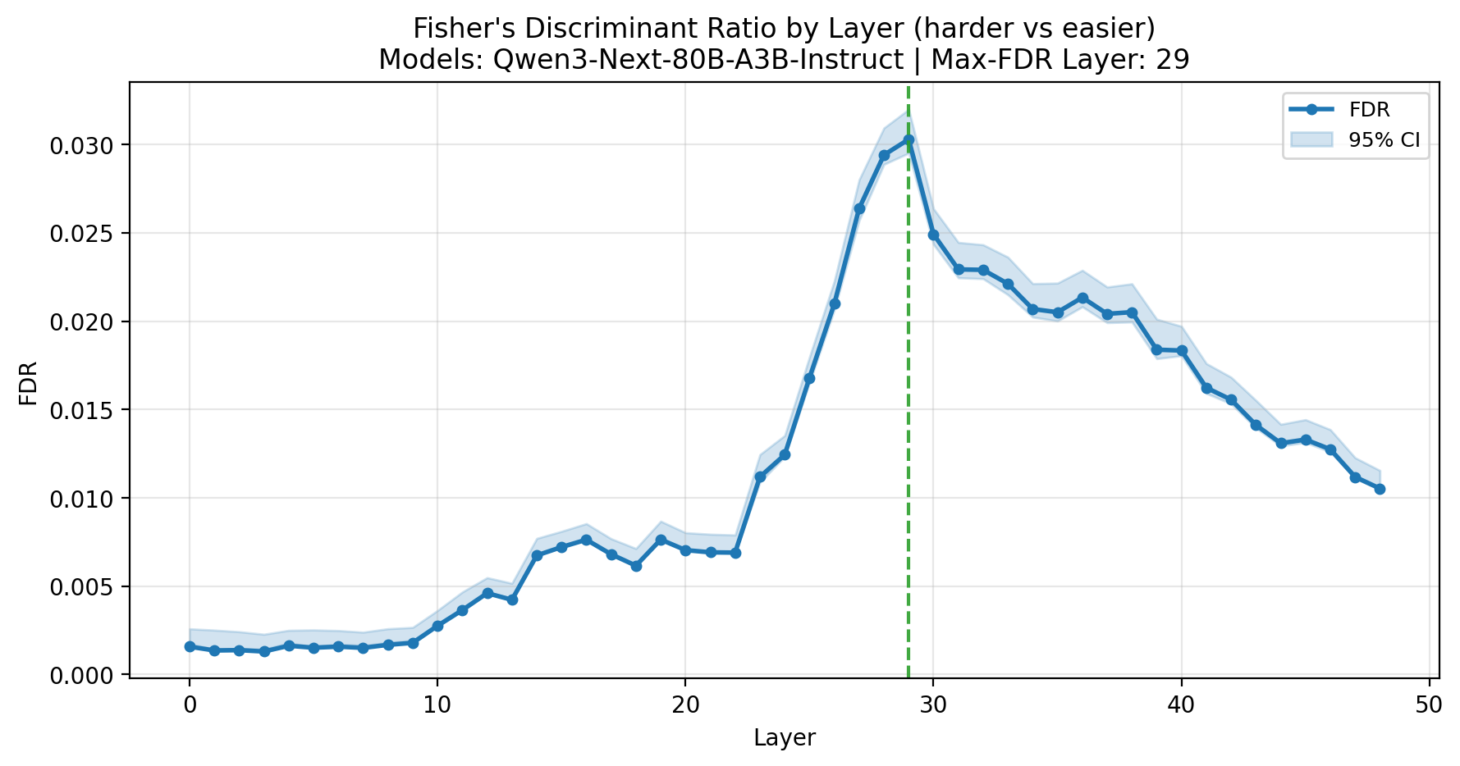}
    \\ \small General Model: Difficulty (Layer 29)
\end{minipage}
\hfill
\begin{minipage}{0.45\textwidth}
    \centering
    \includegraphics[width=\textwidth]{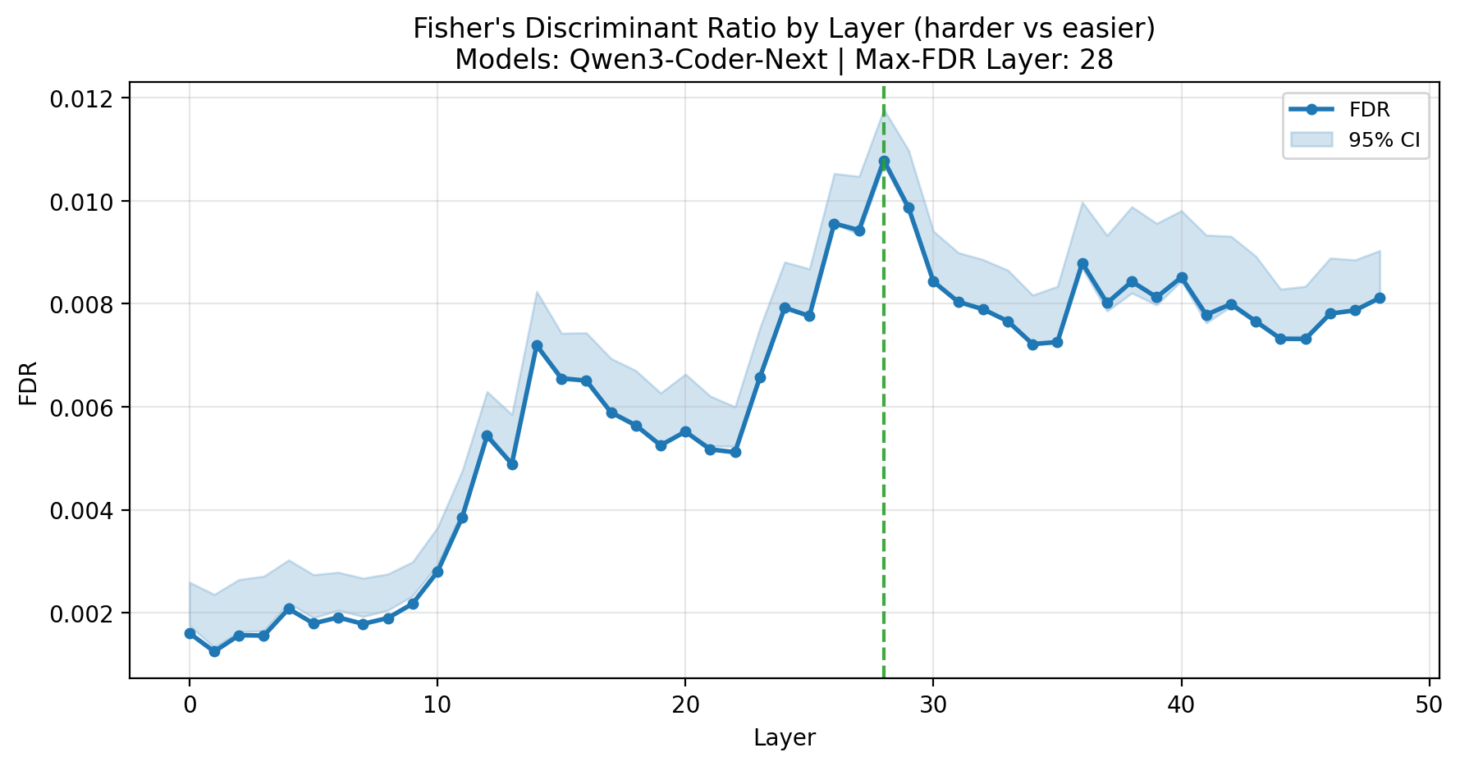}
    \\ \small Coder Model: Difficulty (Layer 28)
\end{minipage}
\\
\vspace{0.5cm}
\begin{minipage}{0.45\textwidth}
    \centering
    \includegraphics[width=\textwidth]{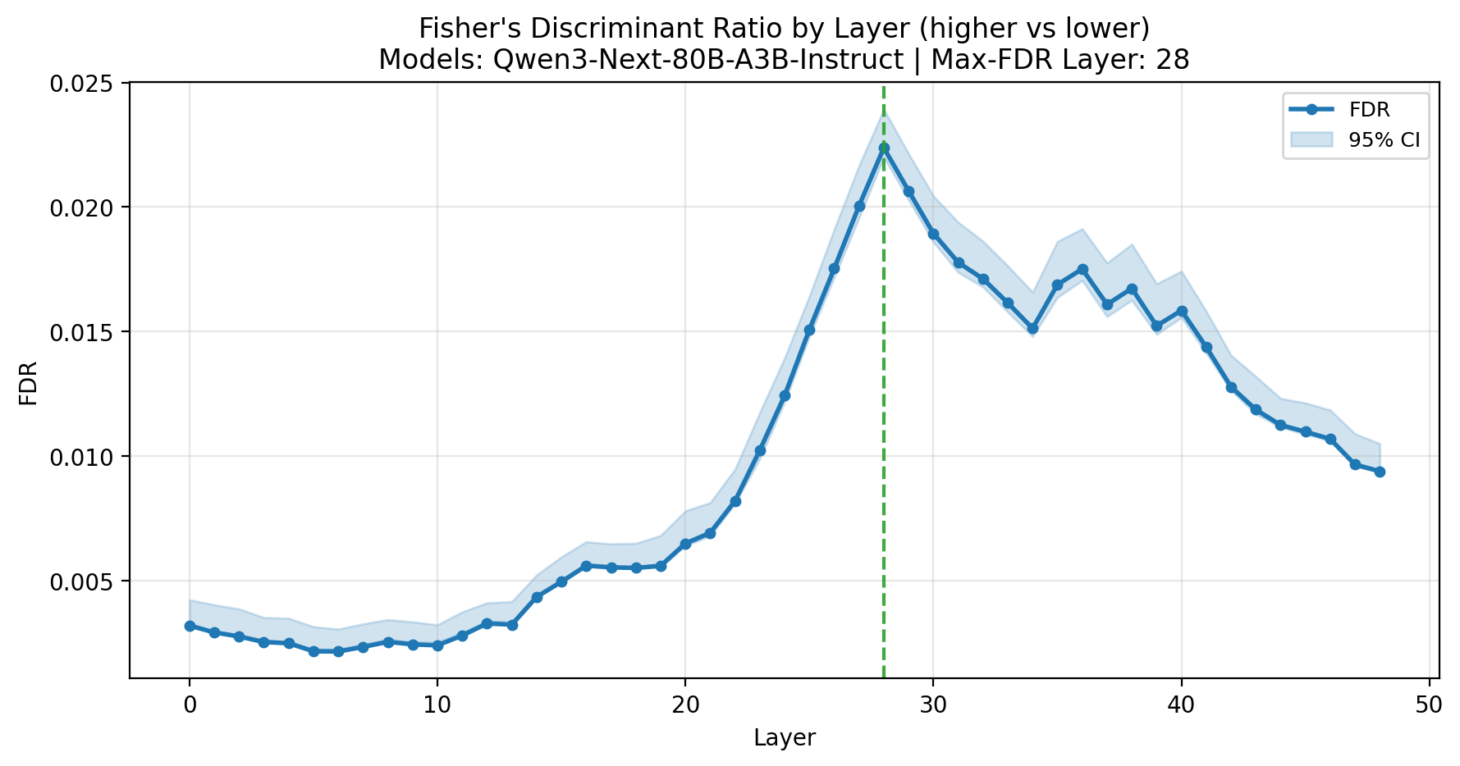}
    \\ \small General Model: Bloom's Levels (Layer 28)
\end{minipage}
\hfill
\begin{minipage}{0.45\textwidth}
    \centering
    \includegraphics[width=\textwidth]{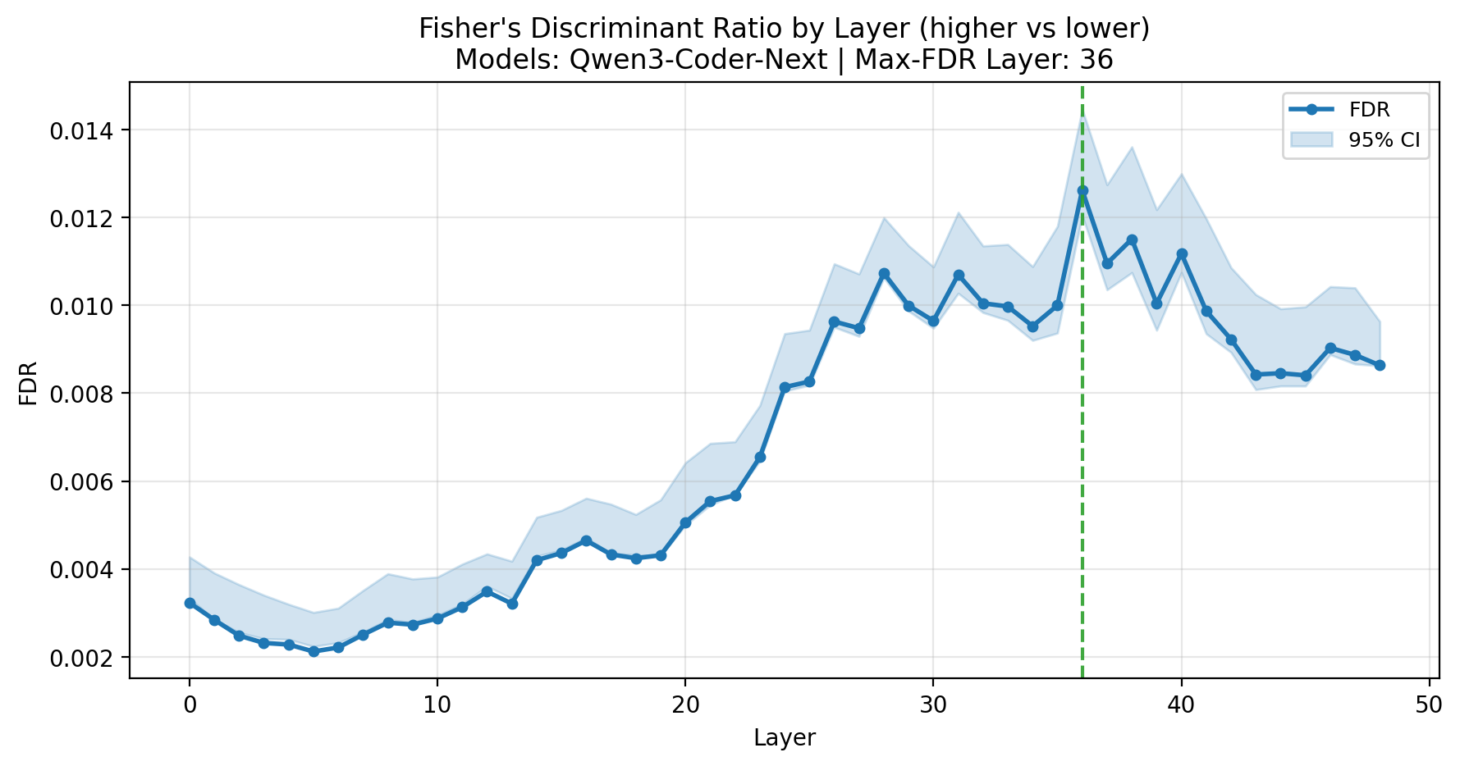}
    \\ \small Coder Model: Bloom's Levels (Layer 36)
\end{minipage}
\caption{Fisher's Discriminant Ratio (FDR) curves showing where difficulty and Bloom-targeted instructions are linearly separable across the 48 hidden layers.}
\label{fig:fisher_curves}
\end{figure}

To locate where these instruction contrasts become most linearly separable internally, we perform layer-by-layer probing using Fisher's Discriminant Ratio (FDR). We calculate 95\% confidence intervals using bootstrap resampling with 1000 samples to verify that the estimated trends are stable. Figure \ref{fig:fisher_curves} shows how this separability, including the confidence bounds, changes across layers for both instruction settings.

The FDR trajectories reveal a difference in separability profiles between the two matched models. For the general model, the peak separability depth for difficulty instructions (``Harder''/``Easier'') and Bloom-target instructions (``Higher''/``Lower'') is nearly identical. Specifically, the model reaches peak linear separability at Layer 29 (FDR $\approx 0.0303$) for difficulty and Layer 28 (FDR $\approx 0.0224$) for Bloom targets. The convergence of these peaks in the middle layers suggests that the two instruction settings are encoded in related regions of the model's residual stream.

In contrast, the coder model exhibits a weaker and deeper separability pattern. For general difficulty instructions, its peak FDR remains low ($\approx 0.0108$ at Layer 28), indicating more overlapping activation distributions under this linear diagnostic. Furthermore, when processing targeted Bloom prompts, the layer of maximum separability shifts deeper into the network, peaking at Layer 36 of 48.

To further inspect the geometric properties of these internal representations, we analyze the residual stream activations using Principal Component Analysis (PCA) in Appendix~\ref{app:pca_embeddings}. These PCA visualizations reveal a dominant direction of variance for both difficulty and Bloom's-target instructions. Leveraging this observation, we compute the mean difference vector ($\mathbf{V}_{\text{diff}}$) between the opposing classes at their peak layers of separability and project the activations onto that axis.

\begin{figure}[ht!]
\centering
\begin{minipage}{0.48\textwidth}
    \centering
    \includegraphics[width=\textwidth]{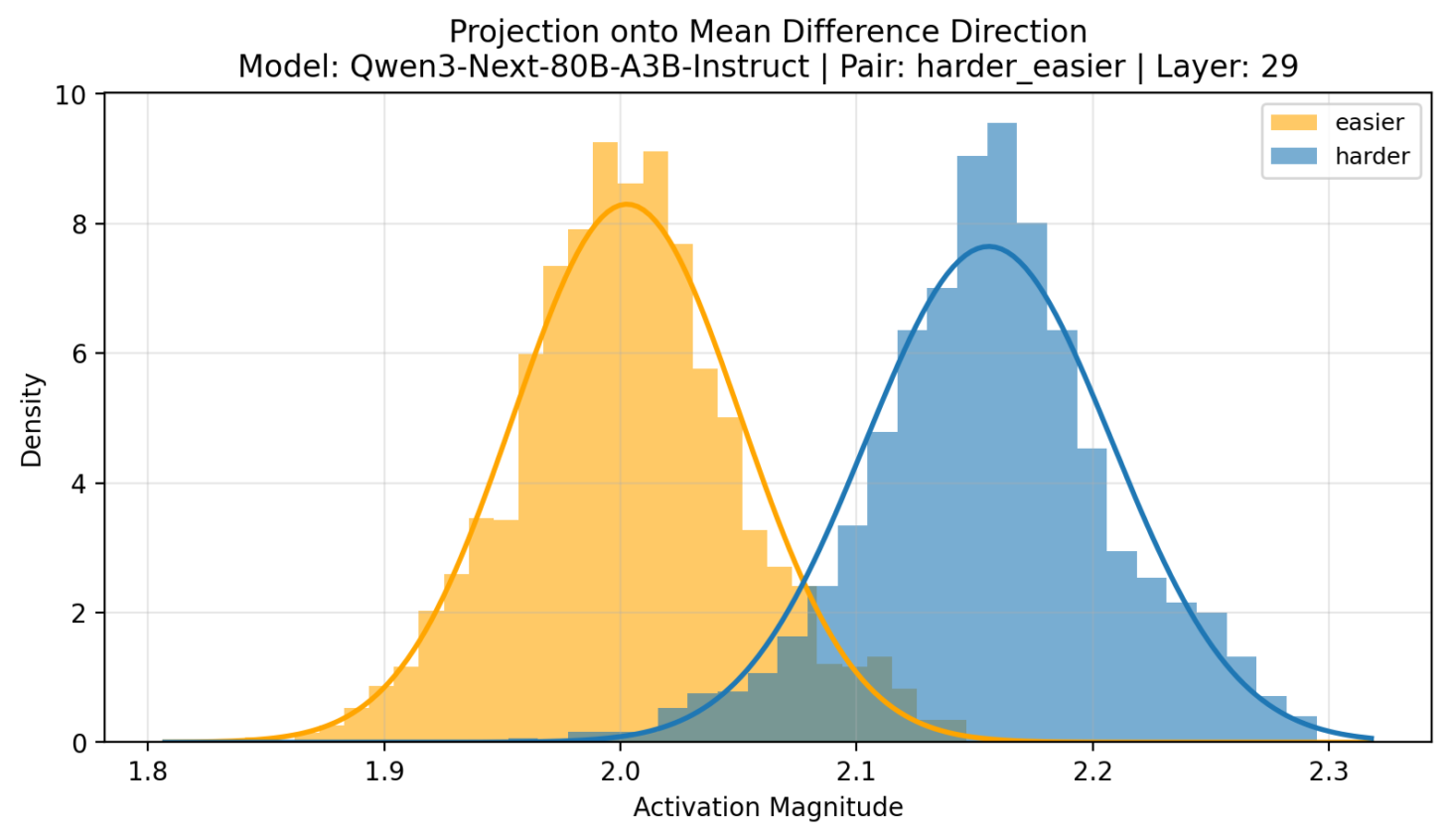}
    \\ \small General Model: Difficulty (Layer 29)
\end{minipage}
\hfill
\begin{minipage}{0.48\textwidth}
    \centering
    \includegraphics[width=\textwidth]{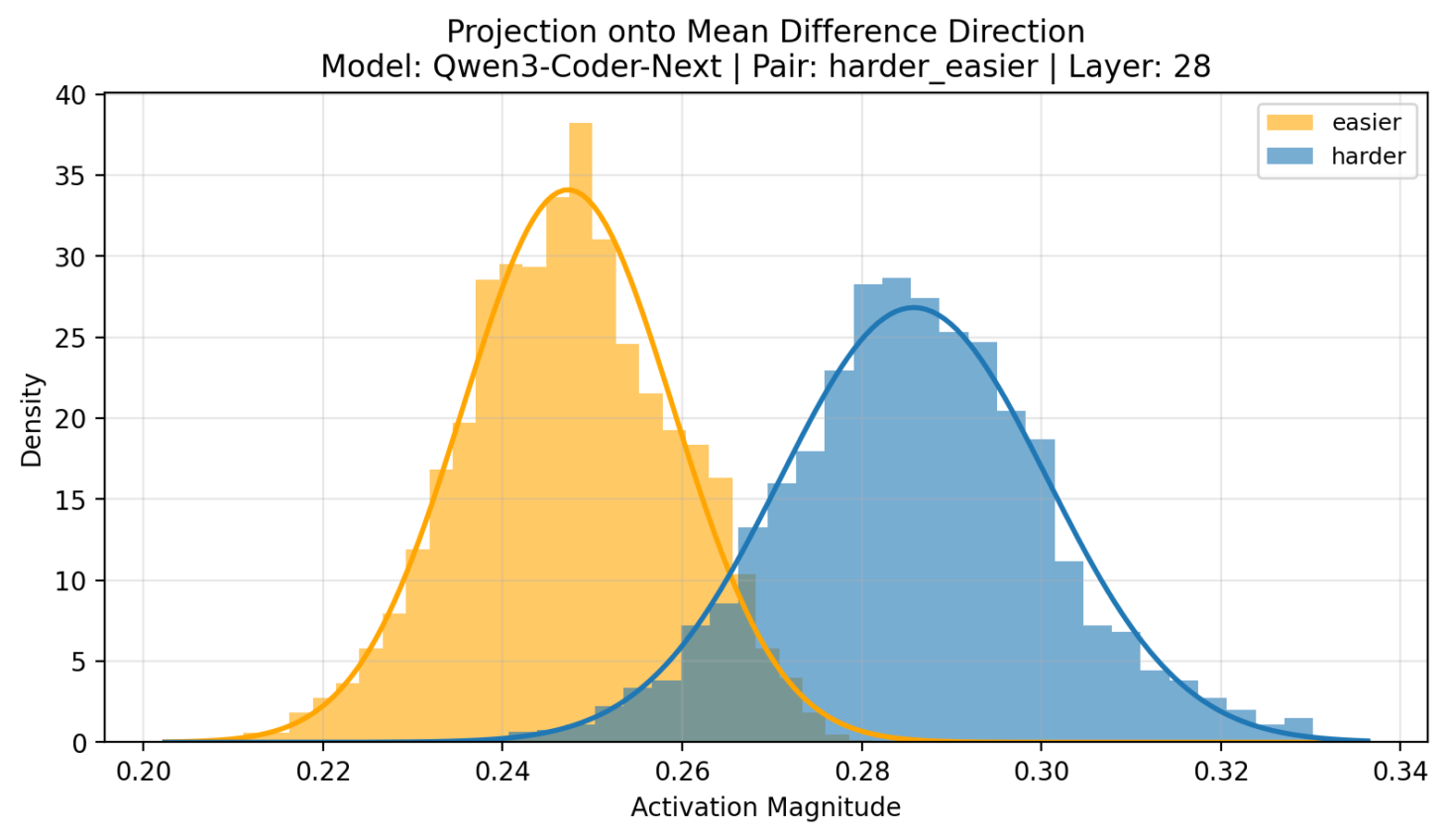}
    \\ \small Coder Model: Difficulty (Layer 28)
\end{minipage}
\caption{Projection of internal activations onto the mean difference direction for difficulty (``Harder'' vs. ``Easier'') at the peak FDR layers.}
\label{fig:latent_projections_difficulty}
\end{figure}

The resulting projection distributions provide a representation-diagnostic counterpart to our earlier behavioral findings. As shown in Figure \ref{fig:latent_projections_difficulty}, the general model displays a more separated distribution between ``Harder'' and ``Easier'' along the projected axis. The coder model shows more overlap and higher variance, which is consistent with its lower FDR score.

\begin{figure}[ht!]
\centering
\begin{minipage}{0.48\textwidth}
    \centering
    \includegraphics[width=\textwidth]{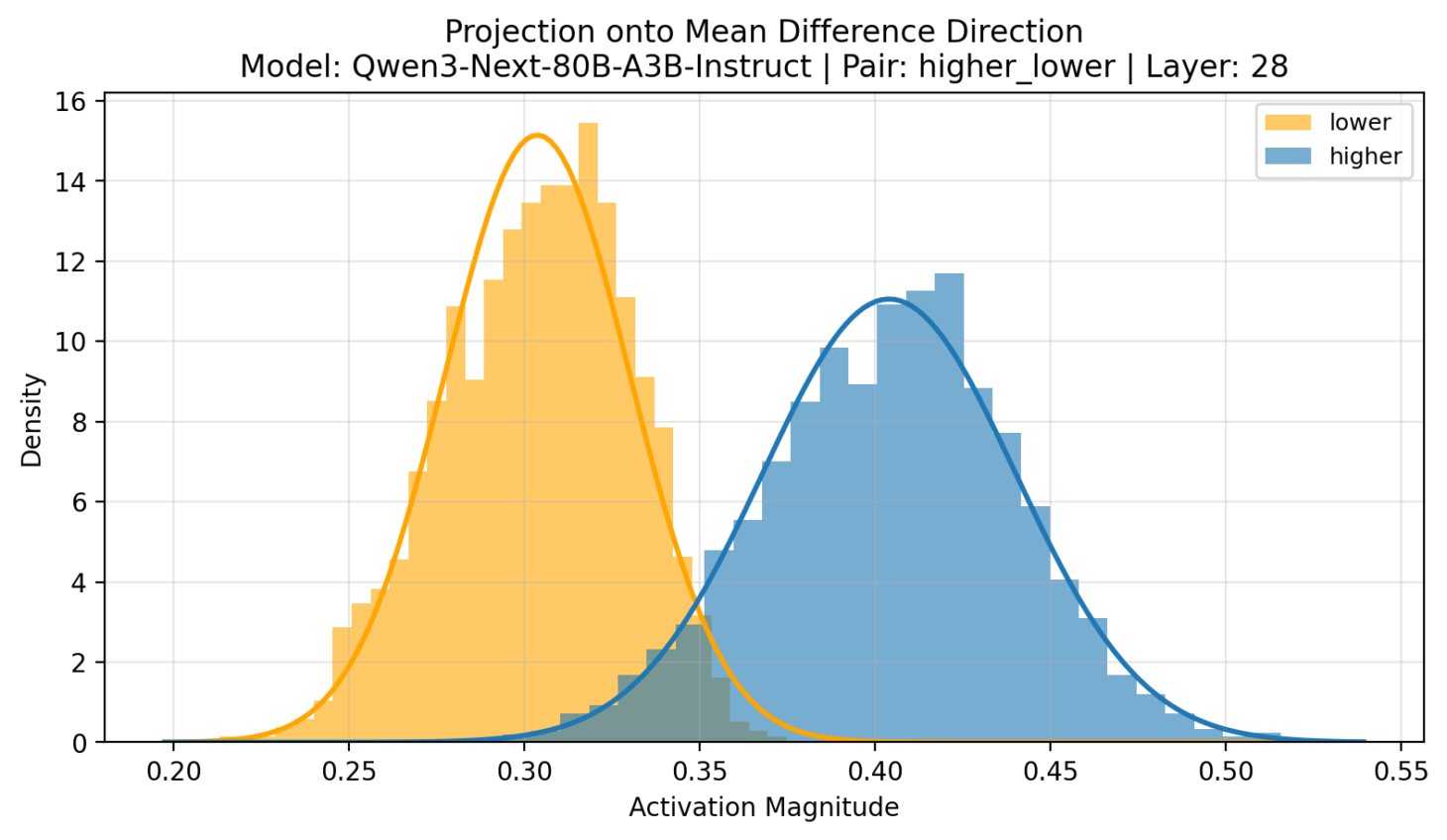}
    \\ \small General Model: Bloom's Levels (Layer 28)
\end{minipage}
\hfill
\begin{minipage}{0.48\textwidth}
    \centering
    \includegraphics[width=\textwidth]{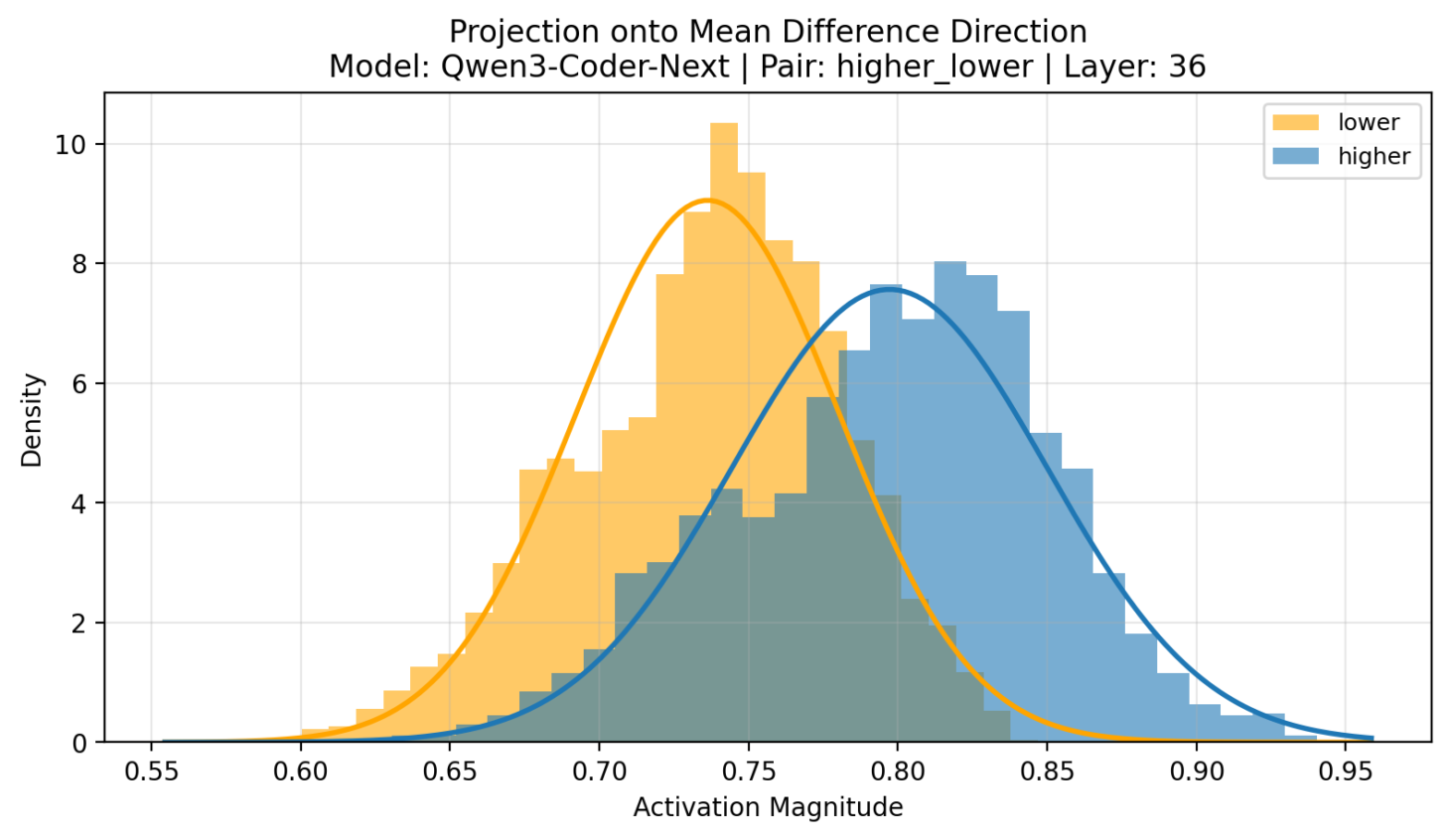}
    \\ \small Coder Model: Bloom's Levels (Layer 36)
\end{minipage}
\caption{Projection of internal activations onto the mean difference direction for targeted Bloom's levels (``Higher'' vs. ``Lower'') at the peak FDR layers.}
\label{fig:latent_projections_bloom}
\end{figure}

This pattern of stronger diagnostic separation in the general model continues in the context of Bloom targets. As illustrated in Figure \ref{fig:latent_projections_bloom}, the general model maintains a sharper projected distribution at its middle-layer peak. Meanwhile, the coder model, which delays this peak separability until Layer 36, still shows substantial overlap between the ``Higher'' and ``Lower'' targets.


\section{Limitations}
\label{sec:limitations}

This study establishes a comparative baseline and highlights several directions for future research. First, our evaluation focuses on two Qwen3-Next models, selected because they provide a closely matched open-weights general/coder comparison with the same architecture and tokenizer at the time of this study. Broader validation across additional model families, scales, and training recipes remains an important next step. Second, using Bloom's Taxonomy as a structural proxy focuses the evaluation on task-level cognitive demand; future work can extend this framework to multi-turn tutoring dynamics, learner-specific adaptation, and alternative pedagogical frameworks. Our evaluation also relies on English-language, Python-centric benchmarks, motivating validation across different programming languages and non-programming learning tasks.

Methodologically, our zero-shot prompting protocol avoids example-copying effects and keeps the intervention directions directly comparable. Future work can evaluate unstructured user instructions, direction-specific prompt tuning, and human audits of generated mutations. Although the Bloom judge is grounded in prior human-alignment validation, generated task mutations may introduce distribution shift. Finally, our representation-diagnostic findings identify where instruction contrasts become most linearly separable; testing the causal role of specific circuits or steering directions requires causal interventions.

\section{Discussion and Conclusion}
\label{sec:discussion}

When applying Large Language Models in education, we must look beyond execution or answer correctness to ask whether they can propose meaningful tasks that align with established learning objectives. The Bloom-aligned framework introduced here provides a concrete way to evaluate whether models can adapt tasks toward intended cognitive demands while preserving instructional intent. Within this controlled model pair, the code-specialized model tends to realize upward shifts through implementation constraints and behavioral inspection, whereas the general model more often reframes tasks into broader design or strategy requirements.

The strongest behavioral pattern surfaced by the framework is a pervasive upward asymmetry: while the models readily synthesize additional cognitive complexity, they struggle to produce lower-level mutations for novice-oriented learning. To serve as effective intelligent tutors, systems must dynamically adjust tasks to match a learner's zone of proximal development. Future work can build on our representation diagnostics by testing whether the mean-difference directions identified here can serve as steering vectors for task-level control. To conclude, when deploying LLMs as educational tools, we should look beyond valid problem-solving and actively assess whether a model can recognize educational frameworks, adjust cognitive complexity, and intentionally mutate tasks for different learners.

\bibliography{colm2026_conference}

@misc{alainUnderstandingIntermediateLayers2018,
  title = {Understanding Intermediate Layers Using Linear Classifier Probes},
  author = {Alain, Guillaume and Bengio, Yoshua},
  year = 2018,
  month = nov,
  number = {arXiv:1610.01644},
  eprint = {1610.01644},
  primaryclass = {stat},
  publisher = {arXiv},
  doi = {10.48550/arXiv.1610.01644},

  archiveprefix = {arXiv},

}

@book{andersonTaxonomyLearningTeaching2001,
  title = {A Taxonomy for Learning, Teaching, and Assessing: {{A}} Revision of {{Bloom}}'s Taxonomy of Educational Objectives: Complete Edition},
  author = {Anderson, Lorin W and Krathwohl, David R},
  year = 2001,
  publisher = {Addison Wesley Longman, Inc.}
}

@misc{anthropicClaude35Haiku2024,
  title = {Claude 3.5 Haiku},
  author = {{Anthropic}},
  year = 2024
}

@misc{austinProgramSynthesisLarge2021,
  title = {Program {{Synthesis}} with {{Large Language Models}}},
  author = {Austin, Jacob and Odena, Augustus and Nye, Maxwell and Bosma, Maarten and Michalewski, Henryk and Dohan, David and Jiang, Ellen and Cai, Carrie and Terry, Michael and Le, Quoc and Sutton, Charles},
  year = 2021,
  month = aug,
  number = {arXiv:2108.07732},
  eprint = {2108.07732},
  primaryclass = {cs},
  publisher = {arXiv},
  doi = {10.48550/arXiv.2108.07732},

  archiveprefix = {arXiv},

}

@article{belinkovProbingClassifiersPromises2022,
  title = {Probing {{Classifiers}}: {{Promises}}, {{Shortcomings}}, and {{Advances}}},
  shorttitle = {Probing {{Classifiers}}},
  author = {Belinkov, Yonatan},
  year = 2022,
  month = apr,
  journal = {Computational Linguistics},
  volume = {48},
  number = {1},
  pages = {207--219},
  issn = {0891-2017, 1530-9312},
  doi = {10.1162/coli_a_00422},

  langid = {english},

}

@book{biggsEvaluatingQualityLearning1982,
  title = {Evaluating the Quality of Learning: The {{SOLO}} Taxonomy (Structure of the Observed Learning Outcome)},
  shorttitle = {Evaluating the Quality of Learning},
  author = {Biggs, John Burville and Biggs, John Burville and Collis, Kevin Francis},
  year = 1982,
  series = {Educational Psychology},
  publisher = {Academic Press},
  address = {New York},
  isbn = {978-0-12-097552-5},
  langid = {english},
  lccn = {370.152 3}
}

@article{brickenMonosemanticityDecomposingLanguage2023,
  title = {Towards Monosemanticity: {{Decomposing}} Language Models with Dictionary Learning},
  author = {Bricken, Trenton and Templeton, Adly and Batson, Joshua and Chen, Brian and Jermyn, Adam and Conerly, Tom and Turner, Nicholas and Anil, Cem and Denison, Carson and Askell, Amanda and Lasenby, Robert and Wu, Yuhuai and Kravec, Shauna and Schiefer, Nicholas and Maxwell, Todd and Joseph, Nicholas and {Hatfield-Dodds}, Zac and Tamkin, Alex and Nguyen, Khanh and McLean, Ben and Burke, Josiah E and Hume, Tristan and Carter, Shan and Henighan, Tom and Olah, Christopher},
  year = 2023,
  journal = {Transformer Circuits Thread}
}

@incollection{campelloDensityBasedClusteringBased2013,
  title = {Density-{{Based Clustering Based}} on {{Hierarchical Density Estimates}}},
  booktitle = {Advances in {{Knowledge Discovery}} and {{Data Mining}}},
  author = {Campello, Ricardo J. G. B. and Moulavi, Davoud and Sander, Joerg},
  editor = {Hutchison, David and Kanade, Takeo and Kittler, Josef and Kleinberg, Jon M. and Mattern, Friedemann and Mitchell, John C. and Naor, Moni and Nierstrasz, Oscar and Pandu Rangan, C. and Steffen, Bernhard and Sudan, Madhu and Terzopoulos, Demetri and Tygar, Doug and Vardi, Moshe Y. and Weikum, Gerhard and Pei, Jian and Tseng, Vincent S. and Cao, Longbing and Motoda, Hiroshi and Xu, Guandong},
  year = 2013,
  volume = {7819},
  pages = {160--172},
  publisher = {Springer Berlin Heidelberg},
  address = {Berlin, Heidelberg},
  doi = {10.1007/978-3-642-37456-2_14},
  isbn = {978-3-642-37455-5 978-3-642-37456-2}
}

@misc{caoQwen3CoderNextTechnicalReport2026,
  title = {Qwen3-{{Coder-Next Technical Report}}},
  author = {Cao, Ruisheng and Chen, Mouxiang and Chen, Jiawei and Cui, Zeyu and Feng, Yunlong and Hui, Binyuan and Jing, Yuheng and Li, Kaixin and Li, Mingze and Lin, Junyang and Ma, Zeyao and Shum, Kashun and Wang, Xuwu and Wei, Jinxi and Yang, Jiaxi and Zhang, Jiajun and Zhang, Lei and Zhang, Zongmeng and Zhao, Wenting and Zhou, Fan},
  year = 2026,
  month = feb,
  number = {arXiv:2603.00729},
  eprint = {2603.00729},
  primaryclass = {cs},
  publisher = {arXiv},
  doi = {10.48550/arXiv.2603.00729},

  archiveprefix = {arXiv},

}

@misc{chenEvaluatingLargeLanguage2021,
  title = {Evaluating {{Large Language Models Trained}} on {{Code}}},
  author = {Chen, Mark and Tworek, Jerry and Jun, Heewoo and Yuan, Qiming and Pinto, Henrique Ponde de Oliveira and Kaplan, Jared and Edwards, Harri and Burda, Yuri and Joseph, Nicholas and Brockman, Greg and Ray, Alex and Puri, Raul and Krueger, Gretchen and Petrov, Michael and Khlaaf, Heidy and Sastry, Girish and Mishkin, Pamela and Chan, Brooke and Gray, Scott and Ryder, Nick and Pavlov, Mikhail and Power, Alethea and Kaiser, Lukasz and Bavarian, Mohammad and Winter, Clemens and Tillet, Philippe and Such, Felipe Petroski and Cummings, Dave and Plappert, Matthias and Chantzis, Fotios and Barnes, Elizabeth and {Herbert-Voss}, Ariel and Guss, William Hebgen and Nichol, Alex and Paino, Alex and Tezak, Nikolas and Tang, Jie and Babuschkin, Igor and Balaji, Suchir and Jain, Shantanu and Saunders, William and Hesse, Christopher and Carr, Andrew N. and Leike, Jan and Achiam, Josh and Misra, Vedant and Morikawa, Evan and Radford, Alec and Knight, Matthew and Brundage, Miles and Murati, Mira and Mayer, Katie and Welinder, Peter and McGrew, Bob and Amodei, Dario and McCandlish, Sam and Sutskever, Ilya and Zaremba, Wojciech},
  year = 2021,
  month = jul,
  number = {arXiv:2107.03374},
  eprint = {2107.03374},
  primaryclass = {cs},
  publisher = {arXiv},
  doi = {10.48550/arXiv.2107.03374},

  archiveprefix = {arXiv},

}

@article{elazarAmnesicProbingBehavioral2021,
  title = {Amnesic {{Probing}}: {{Behavioral Explanation}} with {{Amnesic Counterfactuals}}},
  shorttitle = {Amnesic {{Probing}}},
  author = {Elazar, Yanai and Ravfogel, Shauli and Jacovi, Alon and Goldberg, Yoav},
  year = 2021,
  month = mar,
  journal = {Transactions of the Association for Computational Linguistics},
  volume = {9},
  pages = {160--175},
  issn = {2307-387X},
  doi = {10.1162/tacl_a_00359},

  langid = {english},

}

@article{elhage2021mathematical,
  title = {A Mathematical Framework for Transformer Circuits},
  author = {Elhage, Nelson and Nanda, Neel and Olsson, Catherine and Henighan, Tom and Joseph, Nicholas and Mann, Ben and Askell, Amanda and Bai, Yuntao and Chen, Anna and Conerly, Tom and DasSarma, Nova and Drain, Dawn and Ganguli, Deep and {Hatfield-Dodds}, Zac and Hernandez, Danny and Jones, Andy and Kernion, Jackson and Lovitt, Liane and Ndousse, Kamal and Amodei, Dario and Brown, Tom and Clark, Jack and Kaplan, Jared and McCandlish, Sam and Olah, Chris},
  year = 2021,
  journal = {Transformer Circuits Thread}
}

@misc{grootendorstBERTopicNeuralTopic2022,
  title = {{{BERTopic}}: {{Neural}} Topic Modeling with a Class-Based {{TF-IDF}} Procedure},
  shorttitle = {{{BERTopic}}},
  author = {Grootendorst, Maarten},
  year = 2022,
  month = mar,
  number = {arXiv:2203.05794},
  eprint = {2203.05794},
  primaryclass = {cs},
  publisher = {arXiv},
  doi = {10.48550/arXiv.2203.05794},

  archiveprefix = {arXiv},

}

@misc{guptaHowLLMsUse2026,
  title = {How {{Do LLMs Use Their Depth}}?},
  author = {Gupta, Akshat and Yeung, Jay and Anumanchipalli, Gopala and Ivanova, Anna},
  year = 2026,
  month = mar,
  number = {arXiv:2510.18871},
  eprint = {2510.18871},
  primaryclass = {cs},
  publisher = {arXiv},
  doi = {10.48550/arXiv.2510.18871},

  archiveprefix = {arXiv},

}

@inproceedings{gurneeLanguageModelsRepresent2024,
  title = {Language Models Represent Space and Time},
  booktitle = {International Conference on Learning Representations},
  author = {Gurnee, Wes and Tegmark, Max},
  editor = {Kim, B. and Yue, Y. and Chaudhuri, S. and Fragkiadaki, K. and Khan, M. and Sun, Y.},
  year = 2024,
  volume = {2024},
  pages = {2483--2503}
}

@misc{hkudsDeepTutorAIpoweredPersonalized2025,
  title = {{{DeepTutor}}: {{AI-powered}} Personalized Learning Assistant},
  author = {{HKUDS}},
  year = 2025,
  howpublished = {GitHub}
}

@inproceedings{houEffectsGenerativeAI2024,
  title = {The {{Effects}} of {{Generative AI}} on {{Computing Students}}' {{Help-Seeking Preferences}}},
  booktitle = {Proceedings of the 26th {{Australasian Computing Education Conference}}},
  author = {Hou, Irene and Mettille, Sophia and Man, Owen and Li, Zhuo and Zastudil, Cynthia and MacNeil, Stephen},
  year = 2024,
  month = jan,
  pages = {39--48},
  publisher = {ACM},
  address = {Sydney NSW Australia},
  doi = {10.1145/3636243.3636248},
  isbn = {979-8-4007-1619-5},
  langid = {english}
}

@inproceedings{huberLLMsMeetBloom`s2025,
  title = {{{LLMs}} Meet {{Bloom}}`s {{Taxonomy}}: {{A Cognitive View}} on {{Large Language Model Evaluations}}},
  booktitle = {Proceedings of the 31st {{International Conference}} on {{Computational Linguistics}}},
  author = {Huber, Thomas and Niklaus, Christina},
  editor = {Rambow, Owen and Wanner, Leo and Apidianaki, Marianna and {Al-Khalifa}, Hend and Eugenio, Barbara Di and Schockaert, Steven},
  year = 2025,
  month = jan,
  pages = {5211--5246},
  publisher = {Association for Computational Linguistics},
  address = {Abu Dhabi, UAE},

}

@misc{jainLiveCodeBenchHolisticContamination2024,
  title = {{{LiveCodeBench}}: {{Holistic}} and {{Contamination Free Evaluation}} of {{Large Language Models}} for {{Code}}},
  shorttitle = {{{LiveCodeBench}}},
  author = {Jain, Naman and Han, King and Gu, Alex and Li, Wen-Ding and Yan, Fanjia and Zhang, Tianjun and Wang, Sida and {Solar-Lezama}, Armando and Sen, Koushik and Stoica, Ion},
  year = 2024,
  month = jun,
  number = {arXiv:2403.07974},
  eprint = {2403.07974},
  primaryclass = {cs},
  publisher = {arXiv},
  doi = {10.48550/arXiv.2403.07974},

  archiveprefix = {arXiv},

}

@misc{jangraDesigningEvaluatingChainofHints2026,
  title = {Designing and {{Evaluating Chain-of-Hints}} for {{Scientific Question Answering}}},
  author = {Jangra, Anubhav and Muresan, Smaranda},
  year = 2026,
  month = feb,
  number = {arXiv:2510.21087},
  eprint = {2510.21087},
  primaryclass = {cs},
  publisher = {arXiv},
  doi = {10.48550/arXiv.2510.21087},

  archiveprefix = {arXiv},

}

@inproceedings{jimenezSWEbenchCanLanguage2024,
  title = {{{SWE-bench}}: {{Can Language Models Resolve Real-world Github Issues}}?},
  booktitle = {The {{Twelfth International Conference}} on {{Learning Representations}}},
  author = {Jimenez, Carlos E. and Yang, John and Wettig, Alexander and Yao, Shunyu and Pei, Kexin and Press, Ofir and Narasimhan, Karthik R.},
  year = 2024
}

@inproceedings{kazemitabaarExploringDesignSpace2025,
  title = {Exploring the {{Design Space}} of {{Cognitive Engagement Techniques}} with {{AI-Generated Code}} for {{Enhanced Learning}}},
  booktitle = {Proceedings of the 30th {{International Conference}} on {{Intelligent User Interfaces}}},
  author = {Kazemitabaar, Majeed and Huang, Oliver and Suh, Sangho and Henley, Austin Z and Grossman, Tovi},
  year = 2025,
  month = mar,
  pages = {695--714},
  publisher = {ACM},
  address = {Cagliari Italy},
  doi = {10.1145/3708359.3712104},
  isbn = {979-8-4007-1306-4},
  langid = {english},

}

@inproceedings{liLlmsBuildWorld2024,
  title = {Do Llms Build World Representations? {{Probing}} through the Lens of State Abstraction},
  booktitle = {Advances in Neural Information Processing Systems},
  author = {Li, Zichao and Cao, Yanshuai and Cheung, Jackie C.K.},
  editor = {Globerson, A. and Mackey, L. and Belgrave, D. and Fan, A. and Paquet, U. and Tomczak, J. and Zhang, C.},
  year = 2024,
  volume = {37},
  pages = {98009--98032},
  publisher = {Curran Associates, Inc.},
  doi = {10.52202/079017-3110}
}

@misc{maBloomAPRBloomsTaxonomybased2025,
  title = {{{BloomAPR}}: {{A Bloom}}'s {{Taxonomy-based Framework}} for {{Assessing}} the {{Capabilities}} of {{LLM-Powered APR Solutions}}},
  shorttitle = {{{BloomAPR}}},
  author = {Ma, Yinghang and Shin, Jiho and Silva, Leuson Da and Ming, Zhen and Jiang and Wang, Song and Khomh, Foutse and Tan, Shin Hwei},
  year = 2025,
  month = sep,
  number = {arXiv:2509.25465},
  eprint = {2509.25465},
  primaryclass = {cs},
  publisher = {arXiv},
  doi = {10.48550/arXiv.2509.25465},

  archiveprefix = {arXiv},

}

@inproceedings{marksSparseFeatureCircuits2025,
  title = {Sparse Feature Circuits: {{Discovering}} and Editing Interpretable Causal Graphs in Language Models},
  booktitle = {The Thirteenth International Conference on Learning Representations},
  author = {Marks, Samuel and Rager, Can and Michaud, Eric J and Belinkov, Yonatan and Bau, David and Mueller, Aaron},
  year = 2025
}

@misc{mcinnesUMAPUniformManifold2020,
  title = {{{UMAP}}: {{Uniform Manifold Approximation}} and {{Projection}} for {{Dimension Reduction}}},
  shorttitle = {{{UMAP}}},
  author = {McInnes, Leland and Healy, John and Melville, James},
  year = 2020,
  month = sep,
  number = {arXiv:1802.03426},
  eprint = {1802.03426},
  primaryclass = {stat},
  publisher = {arXiv},
  doi = {10.48550/arXiv.1802.03426},

  archiveprefix = {arXiv},

}

@inproceedings{mengLocatingEditingFactual2022,
  title = {Locating and Editing Factual Associations in {{GPT}}},
  booktitle = {Proceedings of the 36th International Conference on Neural Information Processing Systems},
  author = {Meng, Kevin and Bau, David and Andonian, Alex and Belinkov, Yonatan},
  year = 2022,
  series = {Nips '22},
  publisher = {Curran Associates Inc.},
  address = {Red Hook, NY, USA},

  articleno = {1262},
  isbn = {978-1-7138-7108-8}
}

@article{monroeFightinWordsLexical2008,
  title = {Fightin' {{Words}}: {{Lexical Feature Selection}} and {{Evaluation}} for {{Identifying}} the {{Content}} of {{Political Conflict}}},
  shorttitle = {Fightin' {{Words}}},
  author = {Monroe, Burt L. and Colaresi, Michael P. and Quinn, Kevin M.},
  year = 2008,
  journal = {Political Analysis},
  volume = {16},
  number = {4},
  pages = {372--403},
  issn = {1047-1987, 1476-4989},
  doi = {10.1093/pan/mpn018},

  copyright = {https://www.cambridge.org/core/terms},
  langid = {english},

}

@misc{nostalgebraistInterpretingGPTLogit2020,
  title = {Interpreting {{GPT}}: The Logit Lens},
  author = {{nostalgebraist}},
  year = 2020,
  month = aug
}

@misc{olssonIncontextLearningInduction2022,
  title = {In-Context {{Learning}} and {{Induction Heads}}},
  author = {Olsson, Catherine and Elhage, Nelson and Nanda, Neel and Joseph, Nicholas and DasSarma, Nova and Henighan, Tom and Mann, Ben and Askell, Amanda and Bai, Yuntao and Chen, Anna and Conerly, Tom and Drain, Dawn and Ganguli, Deep and {Hatfield-Dodds}, Zac and Hernandez, Danny and Johnston, Scott and Jones, Andy and Kernion, Jackson and Lovitt, Liane and Ndousse, Kamal and Amodei, Dario and Brown, Tom and Clark, Jack and Kaplan, Jared and McCandlish, Sam and Olah, Chris},
  year = 2022,
  month = sep,
  number = {arXiv:2209.11895},
  eprint = {2209.11895},
  primaryclass = {cs},
  publisher = {arXiv},
  doi = {10.48550/arXiv.2209.11895},

  archiveprefix = {arXiv},

}

@misc{openaiIntroducingSWEbenchVerified2024,
  title = {Introducing {{SWE-bench Verified}}},
  author = {{OpenAI}},
  year = 2024
}

@inproceedings{parkLinearRepresentationHypothesis2024,
  title = {The Linear Representation Hypothesis and the Geometry of Large Language Models},
  booktitle = {Proceedings of the 41st International Conference on Machine Learning},
  author = {Park, Kiho and Choe, Yo Joong and Veitch, Victor},
  editor = {Salakhutdinov, Ruslan and Kolter, Zico and Heller, Katherine and Weller, Adrian and Oliver, Nuria and Scarlett, Jonathan and Berkenkamp, Felix},
  year = 2024,
  month = jul,
  series = {Proceedings of Machine Learning Research},
  volume = {235},
  pages = {39643--39666},
  publisher = {PMLR},

}

@inproceedings{pratherWideningGapBenefits2024,
  title = {The {{Widening Gap}}: {{The Benefits}} and {{Harms}} of {{Generative AI}} for {{Novice Programmers}}},
  shorttitle = {The {{Widening Gap}}},
  booktitle = {Proceedings of the 2024 {{ACM Conference}} on {{International Computing Education Research}} - {{Volume}} 1},
  author = {Prather, James and Reeves, Brent N and Leinonen, Juho and MacNeil, Stephen and Randrianasolo, Arisoa S and Becker, Brett A. and Kimmel, Bailey and Wright, Jared and Briggs, Ben},
  year = 2024,
  month = aug,
  pages = {469--486},
  publisher = {ACM},
  address = {Melbourne VIC Australia},
  doi = {10.1145/3632620.3671116},
  isbn = {979-8-4007-0475-8},
  langid = {english}
}

@misc{raimondiMechanisticInterpretabilityCognitive2026,
  title = {Mechanistic {{Interpretability}} of {{Cognitive Complexity}} in {{LLMs}} via {{Linear Probing}} Using {{Bloom}}'s {{Taxonomy}}},
  author = {Raimondi, Bianca and Gabbrielli, Maurizio},
  year = 2026,
  month = feb,
  number = {arXiv:2602.17229},
  eprint = {2602.17229},
  primaryclass = {cs},
  publisher = {arXiv},
  doi = {10.48550/arXiv.2602.17229},

  archiveprefix = {arXiv},

}

@inproceedings{ravichanderProbingProbingParadigm2021,
  title = {Probing the {{Probing Paradigm}}: {{Does Probing Accuracy Entail Task Relevance}}?},
  shorttitle = {Probing the {{Probing Paradigm}}},
  booktitle = {Proceedings of the 16th {{Conference}} of the {{European Chapter}} of the {{Association}} for {{Computational Linguistics}}: {{Main Volume}}},
  author = {Ravichander, Abhilasha and Belinkov, Yonatan and Hovy, Eduard},
  year = 2021,
  pages = {3363--3377},
  publisher = {Association for Computational Linguistics},
  address = {Online},
  doi = {10.18653/v1/2021.eacl-main.295},
  langid = {english},

}

@misc{reimersSentenceBERTSentenceEmbeddings2019,
  title = {Sentence-{{BERT}}: {{Sentence Embeddings}} Using {{Siamese BERT-Networks}}},
  shorttitle = {Sentence-{{BERT}}},
  author = {Reimers, Nils and Gurevych, Iryna},
  year = 2019,
  month = aug,
  number = {arXiv:1908.10084},
  eprint = {1908.10084},
  primaryclass = {cs},
  publisher = {arXiv},
  doi = {10.48550/arXiv.1908.10084},

  archiveprefix = {arXiv},

}

@inproceedings{skeanLayerLayerUncovering2025,
  title = {Layer by Layer: {{Uncovering}} Hidden Representations in Language Models},
  booktitle = {Forty-Second International Conference on Machine Learning},
  author = {Skean, Oscar and Arefin, Md Rifat and Zhao, Dan and Patel, Niket Nikul and Naghiyev, Jalal and LeCun, Yann and {Shwartz-Ziv}, Ravid},
  year = 2025
}

@book{vygotskyMindSocietyDevelopment1980,
  title = {Mind in {{Society}}: {{Development}} of {{Higher Psychological Processes}}},
  shorttitle = {Mind in {{Society}}},
  author = {Vygotsky, L. S.},
  editor = {Cole, Michael and {Jolm-Steiner}, Vera and Scribner, Sylvia and Souberman, Ellen},
  year = 1980,
  month = oct,
  eprint = {10.2307/j.ctvjf9vz4},
  eprinttype = {jstor},
  publisher = {Harvard University Press},
  doi = {10.2307/j.ctvjf9vz4},
  isbn = {978-0-674-07668-6 978-0-674-57628-5}
}

@inproceedings{wangInterpretabilityWildCircuit2023,
  title = {Interpretability in the Wild: A Circuit for Indirect Object Identification in {{GPT-2}} Small},
  booktitle = {The Eleventh International Conference on Learning Representations},
  author = {Wang, Kevin Ro and Variengien, Alexandre and Conmy, Arthur and Shlegeris, Buck and Steinhardt, Jacob},
  year = 2023
}

@techreport{webbCriteriaAlignmentExpectations1997,
  title = {Criteria for Alignment of Expectations and Assessments in Mathematics and Science Education. {{Research}} Monograph No. 6},
  author = {Webb, Norman L.},
  year = 1997,
  address = {Washington, DC},
  institution = {Council of Chief State School Officers}
}

@misc{yangQwen3TechnicalReport2025,
  title = {Qwen3 {{Technical Report}}},
  author = {Yang, An and Li, Anfeng and Yang, Baosong and Zhang, Beichen and Hui, Binyuan and Zheng, Bo and Yu, Bowen and Gao, Chang and Huang, Chengen and Lv, Chenxu and Zheng, Chujie and Liu, Dayiheng and Zhou, Fan and Huang, Fei and Hu, Feng and Ge, Hao and Wei, Haoran and Lin, Huan and Tang, Jialong and Yang, Jian and Tu, Jianhong and Zhang, Jianwei and Yang, Jianxin and Yang, Jiaxi and Zhou, Jing and Zhou, Jingren and Lin, Junyang and Dang, Kai and Bao, Keqin and Yang, Kexin and Yu, Le and Deng, Lianghao and Li, Mei and Xue, Mingfeng and Li, Mingze and Zhang, Pei and Wang, Peng and Zhu, Qin and Men, Rui and Gao, Ruize and Liu, Shixuan and Luo, Shuang and Li, Tianhao and Tang, Tianyi and Yin, Wenbiao and Ren, Xingzhang and Wang, Xinyu and Zhang, Xinyu and Ren, Xuancheng and Fan, Yang and Su, Yang and Zhang, Yichang and Zhang, Yinger and Wan, Yu and Liu, Yuqiong and Wang, Zekun and Cui, Zeyu and Zhang, Zhenru and Zhou, Zhipeng and Qiu, Zihan},
  year = 2025,
  month = may,
  number = {arXiv:2505.09388},
  eprint = {2505.09388},
  primaryclass = {cs},
  publisher = {arXiv},
  doi = {10.48550/arXiv.2505.09388},

  archiveprefix = {arXiv},

}

@misc{yinNeuronGuidedInterpretationCode2025,
  title = {Neuron-{{Guided Interpretation}} of {{Code LLMs}}: {{Where}}, {{Why}}, and {{How}}?},
  shorttitle = {Neuron-{{Guided Interpretation}} of {{Code LLMs}}},
  author = {Yin, Zhe and Gu, Xiaodong and Shen, Beijun},
  year = 2025,
  month = dec,
  number = {arXiv:2512.19980},
  eprint = {2512.19980},
  primaryclass = {cs},
  publisher = {arXiv},
  doi = {10.48550/arXiv.2512.19980},

  archiveprefix = {arXiv},

}

@article{yuMOOCMAICReimagine2026,
  title = {From {{MOOC}} to {{MAIC}}: {{Reimagine}} Online Teaching and Learning through {{LLM-driven}} Agents},
  author = {Yu, Ji-Fan and {Zhang-Li}, Daniel and Zhang, Zhe-Yuan and Wang, Yu-Cheng and Li, Hao-Xuan and Lim, Joy Jia Yin and Hao, Zhan-Xin and Tu, Shang-Qing and Zhang, Lu and Dai, Xu-Sheng and Jiang, Jian-Xiao and Yang, Shen and Qin, Fei and Li, Ze-Kun and Cong, Xin and Xu, Bin and Hou, Lei and Li, Man-Li and Li, Juan-Zi and Liu, Hui-Qin and Zhang, Yu and Liu, Zhi-Yuan and Sun, Mao-Song},
  year = 2026,
  journal = {Journal of Computer Science and Technology},
  issn = {1000-9000(Print) /1860-4749(Online)},
  doi = {10.1007/s11390-025-6000-0}
}

@incollection{yuRecallReasoningAutomated2025,
  title = {From {{Recall}} to {{Reasoning}}: {{Automated Question Generation}} for {{Deeper Math Learning Through Large Language Models}}},
  shorttitle = {From {{Recall}} to {{Reasoning}}},
  booktitle = {Artificial {{Intelligence}} in {{Education}}},
  author = {Yu, Yongan and Krantz, Alexandre and Lobczowski, Nikki G.},
  editor = {Cristea, Alexandra I. and Walker, Erin and Lu, Yu and Santos, Olga C. and Isotani, Seiji},
  year = 2025,
  volume = {15881},
  pages = {414--422},
  publisher = {Springer Nature Switzerland},
  address = {Cham},
  doi = {10.1007/978-3-031-98462-4_52},
  isbn = {978-3-031-98461-7 978-3-031-98462-4},
  langid = {english}
}

@inproceedings{zhangLLMBasedFrameworkSimulating2025,
  title = {An {{LLM-Based Framework}} for {{Simulating}}, {{Classifying}}, and {{Correcting Students}}' {{Programming Knowledge}} with the {{SOLO Taxonomy}}},
  booktitle = {Proceedings of the 56th {{ACM Technical Symposium}} on {{Computer Science Education V}}. 2},
  author = {Zhang, Shan and Meshram, Pragati Shuddhodhan and Ganapathy Prasad, Priyadharshini and Israel, Maya and Bhat, Suma},
  year = 2025,
  month = feb,
  pages = {1681--1682},
  publisher = {ACM},
  address = {Pittsburgh PA USA},
  doi = {10.1145/3641555.3705125},
  isbn = {979-8-4007-0532-8},
  langid = {english}
}

@inproceedings{zhuoBigCodeBenchBenchmarkingCode2025,
  title = {{{BigCodeBench}}: {{Benchmarking Code Generation}} with {{Diverse Function Calls}} and {{Complex Instructions}}},
  booktitle = {The {{Thirteenth International Conference}} on {{Learning Representations}}},
  author = {Zhuo, Terry Yue and Chien, Vu Minh and Chim, Jenny and Hu, Han and Yu, Wenhao and Widyasari, Ratnadira and Yusuf, Imam Nur Bani and Zhan, Haolan and He, Junda and Paul, Indraneil and Brunner, Simon and GONG, Chen and Hoang, James and Zebaze, Armel Randy and Hong, Xiaoheng and Li, Wen-Ding and Kaddour, Jean and Xu, Ming and Zhang, Zhihan and Yadav, Prateek and Jain, Naman and Gu, Alex and Cheng, Zhoujun and Liu, Jiawei and Liu, Qian and Wang, Zijian and Lo, David and Hui, Binyuan and Muennighoff, Niklas and Fried, Daniel and Du, Xiaoning and de Vries, Harm and Werra, Leandro Von},
  year = 2025
}

@misc{zoumpoulidiBloomWiseEnhancingProblemSolving2024,
  title = {{{BloomWise}}: {{Enhancing Problem-Solving}} Capabilities of {{Large Language Models}} Using {{Bloom}}'s-{{Taxonomy-Inspired Prompts}}},
  shorttitle = {{{BloomWise}}},
  author = {Zoumpoulidi, Maria-Eleni and Paraskevopoulos, Georgios and Potamianos, Alexandros},
  year = 2024,
  month = oct,
  number = {arXiv:2410.04094},
  eprint = {2410.04094},
  primaryclass = {cs},
  publisher = {arXiv},
  doi = {10.48550/arXiv.2410.04094},

  archiveprefix = {arXiv},

}

@misc{zouRepresentationEngineeringTopDown2025,
  title = {Representation {{Engineering}}: {{A Top-Down Approach}} to {{AI Transparency}}},
  shorttitle = {Representation {{Engineering}}},
  author = {Zou, Andy and Phan, Long and Chen, Sarah and Campbell, James and Guo, Phillip and Ren, Richard and Pan, Alexander and Yin, Xuwang and Mazeika, Mantas and Dombrowski, Ann-Kathrin and Goel, Shashwat and Li, Nathaniel and Byun, Michael J. and Wang, Zifan and Mallen, Alex and Basart, Steven and Koyejo, Sanmi and Song, Dawn and Fredrikson, Matt and Kolter, J. Zico and Hendrycks, Dan},
  year = 2025,
  month = mar,
  number = {arXiv:2310.01405},
  eprint = {2310.01405},
  primaryclass = {cs},
  publisher = {arXiv},
  doi = {10.48550/arXiv.2310.01405},

  archiveprefix = {arXiv},

}

@article{Zhang_Rayz_2026, title={Do Programmers and AI See the Same Problem? Quantifying Cognitive Misalignment in Code Generation}, volume={39}, url={https://journals.flvc.org/FLAIRS/article/view/141770}, DOI={10.32473/flairs.39.1.141770}, abstractNote={&amp;lt;p&amp;gt;The integration of AI assistants into software development raises fundamental questions about how task complexity is evaluated and the extent to which these evaluations align with human perception. Current evaluations focus primarily on functional correctness, overlooking this cognitive alignment. We introduce and empirically examine cognitive misalignment: the discrepancy between human and AI perceptions of a task&amp;#039;s cognitive demands. Using Bloom’s Taxonomy, we prompt five LLMs to classify 2,520 tasks from three code generation benchmarks, and establish human reference annotations for 150 tasks via expert consensus. Results show systematic misalignment: humans predominantly classify tasks as &amp;quot;Apply&amp;quot; or &amp;quot;Analyze&amp;quot;, whereas several LLMs overestimate the &amp;quot;Create&amp;quot; dimension. This gap varies by model and task type and may contribute to observed interaction frictions and productivity paradoxes. Our findings motivate the development of cognitively aware benchmarks and evaluation methods that better reflect human judgments of task complexity.&amp;lt;/p&amp;gt;}, number={1}, journal={The International FLAIRS Conference Proceedings}, author={Zhang, Yi and Rayz, Julia}, year={2026}, month={May} }
\bibliographystyle{colm2026_conference}

\appendix

\clearpage
\section{Generation Parameters and Prompt Templates}
\label{app:settings}

The generation experiments used a fixed decoding configuration so the observed differences could be attributed to the instruction type and model family rather than sampling noise. The prompt settings are summarized below.

\begin{table}[ht!]
\centering
\begin{tabular}{@{}ll@{}}
\toprule
Parameter & Value \\ \midrule
Temperature & 0.1 \\
Top-p & 0.95 \\
Frequency penalty & 0.1 \\
Maximum output tokens & 2048 \\
Random seed & 42 \\
Hardware & Four NVIDIA RTX A6000 GPUs \\
\bottomrule
\end{tabular}
\caption{Generation parameter settings used for the intervention experiments.}
\label{tab:prompt_parameters}
\end{table}

\subsection{Prompt Templates}

The prompt templates used in the experiments are shown below. The first template is the Bloom classification prompt, and the second is the task-intervention template used to generate the four mutation conditions.

\begin{tcolorbox}[title=Bloom's Classification Prompt, fonttitle=\bfseries, colback=gray!5!white, colframe=gray!75!black]
\small
Please analyze the following coding question and determine which level of Bloom's Taxonomy it most closely aligns with. The six levels of Bloom's Taxonomy, from lowest to highest cognitive complexity, are: Remember, Understand, Apply, Analyze, Evaluate, Create.

\medskip
\textbf{Instructions} \\
1. Determine which Bloom's Taxonomy level best matches the skills required to answer this question. \\
2. Provide your reasoning, explaining specific elements of the question that correspond to this level. \\
3. Return ONLY an XML object with tags: \texttt{<level>}, \texttt{<rationale>}. If you include anything else, the response is invalid.

\medskip
\textbf{Format} \\
Please format your response as valid XML with the following structure:
\begin{verbatim}
<result>
  <level>
    One of: Remember, Understand, Apply, Analyze, Evaluate, Create
  </level>
  <rationale>
    Brief explanation of why this question fits the identified level
  </rationale>
</result>
\end{verbatim}
Please ensure your response is valid XML without any additional text or markdown formatting outside the XML structure.

\medskip
\textbf{Coding Question to Analyze} \\
Question: \{question\}
\end{tcolorbox}

\begin{tcolorbox}[title=Shared Output Specifications, fonttitle=\bfseries, colback=gray!5!white, colframe=gray!75!black]
\small
\textbf{Output Specifications} \\
1. MODE: Output your FIRST draft immediately without verification. \\
2. FORMAT: Output strictly a single new question text wrapped in \texttt{<new\_question> ... </new\_question>} tags. \\
3. CONTENT WHITELIST: The output must consist ONLY of the Problem Description and (optionally) Input/Output examples. \\
4. NO ASCII OR STATE TRACES: Do NOT generate ASCII art (grids/graphs) or step-by-step execution logs (e.g., \texttt{State A -> State B}). \\
5. EXAMPLE STRUCTURE: You may include 0, 1, or 2 examples. Each example must consist STRICTLY of \texttt{Input} and \texttt{Output} fields only. No Explanation. \\
6. TERMINATION PROTOCOL: End the response immediately after the last example. Do not add post-scripts or notes.
\end{tcolorbox}

\begin{tcolorbox}[title=Task Intervention Template, fonttitle=\bfseries, colback=gray!5!white, colframe=gray!75!black]
\small
Task: Rewrite the coding question inside \texttt{<original\_question>} to \textbf{[TASK\_GOAL]}.

\small
[SHARED\_SPECS]

\small
\textbf{[BLOOM\_CONTEXT]}

\small
Direction: \textbf{[SPECIFIC\_DIRECTION]}

\medskip
Input: \\
\texttt{<original\_question>} \\
\{question\} \\
\texttt{</original\_question>}

\medskip
Output:
\end{tcolorbox}

\subsection{Intervention Settings}

The generation experiments used a fixed intervention template with four conditions. Two conditions implement general difficulty control, and two conditions implement Bloom's control. The shared template keeps the four intervention directions comparable. The Harder/Easier conditions are observational with respect to Bloom's levels: they show how general difficulty language behaves on the Bloom scale alongside the Higher/Lower Bloom targets. The exact parameterization is summarized below.

\begin{itemize}
\item \textbf{Difficulty (Harder)}
\begin{itemize}
\item \texttt{[TASK\_GOAL]}: make it HARDER
\item \texttt{[BLOOM\_CONTEXT]}: none
\item \texttt{[SPECIFIC\_DIRECTION]}: Increase the difficulty of the original question.
\end{itemize}
\item \textbf{Difficulty (Easier)}
\begin{itemize}
\item \texttt{[TASK\_GOAL]}: make it EASIER
\item \texttt{[BLOOM\_CONTEXT]}: none
\item \texttt{[SPECIFIC\_DIRECTION]}: Reduce the difficulty of the original question.
\end{itemize}
\item \textbf{Bloom's Control (Higher)}
\begin{itemize}
\item \texttt{[TASK\_GOAL]}: target a HIGHER level of Bloom's Taxonomy
\item \texttt{[BLOOM\_CONTEXT]}: Bloom's Levels: Remember, Understand, Apply, Analyze, Evaluate, Create.
\item \texttt{[SPECIFIC\_DIRECTION]}: Target a higher Bloom's level: Evaluate or Create.
\end{itemize}
\item \textbf{Bloom's Control (Lower)}
\begin{itemize}
\item \texttt{[TASK\_GOAL]}: target a LOWER level of Bloom's Taxonomy
\item \texttt{[BLOOM\_CONTEXT]}: Bloom's Levels: Remember, Understand, Apply, Analyze, Evaluate, Create.
\item \texttt{[SPECIFIC\_DIRECTION]}: Target a lower Bloom's level: Remember or Understand.
\end{itemize}
\end{itemize}

\clearpage
\section{Transition Heatmaps}
\label{app:transition}
\begin{figure}[ht!]
\centering
\begin{minipage}{0.48\textwidth}
    \centering
    \includegraphics[width=\textwidth]{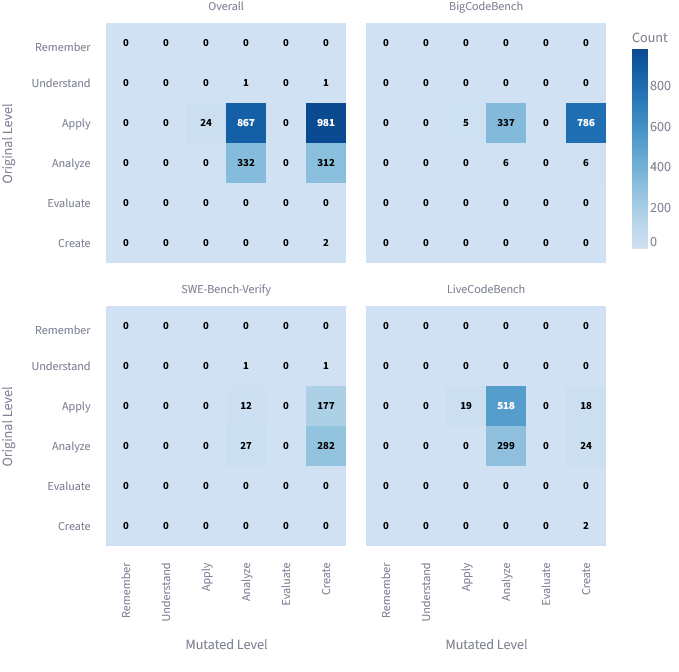}
    \\ \small Harder Intervention
\end{minipage}
\hfill
\begin{minipage}{0.48\textwidth}
    \centering
    \includegraphics[width=\textwidth]{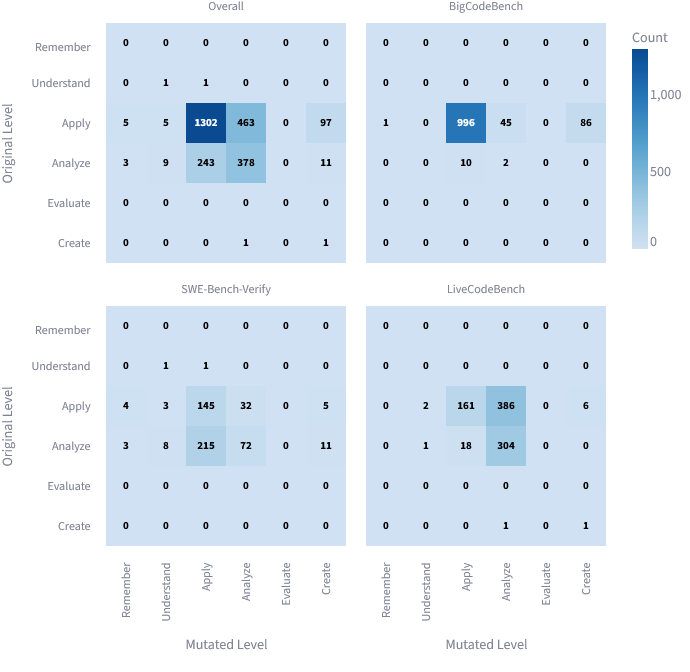}
    \\ \small Easier Intervention
\end{minipage}
\\
\vspace{0.5cm}
\begin{minipage}{0.48\textwidth}
    \centering
    \includegraphics[width=\textwidth]{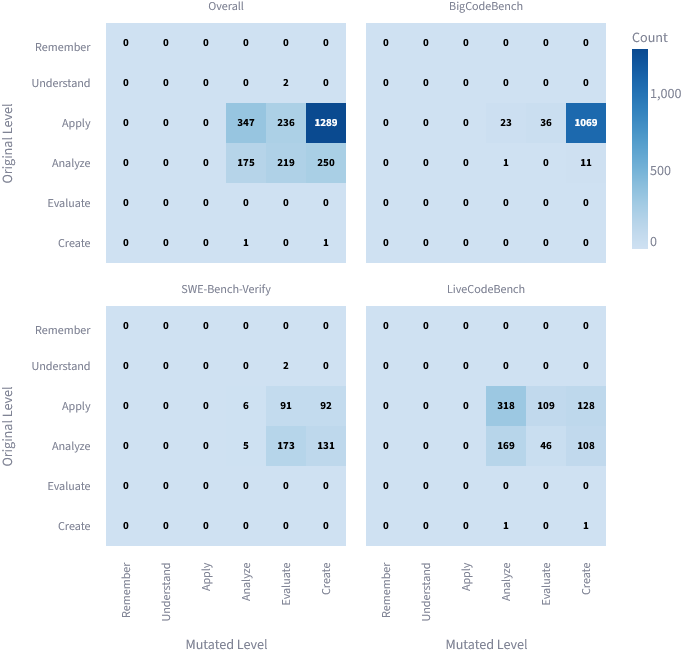}
    \\ \small Higher Target
\end{minipage}
\hfill
\begin{minipage}{0.48\textwidth}
    \centering
    \includegraphics[width=\textwidth]{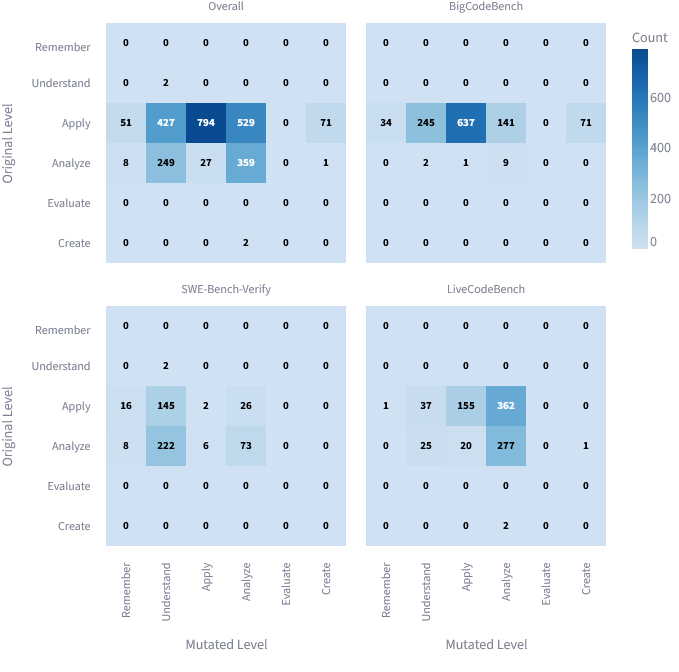}
    \\ \small Lower Target
\end{minipage}
\caption{Transition heatmaps for Qwen3-Next-80B-A3B (General) across different intervention settings.}
\label{fig:transition_instruct}
\end{figure}

\begin{figure}[ht!]
\centering
\begin{minipage}{0.48\textwidth}
    \centering
    \includegraphics[width=\textwidth]{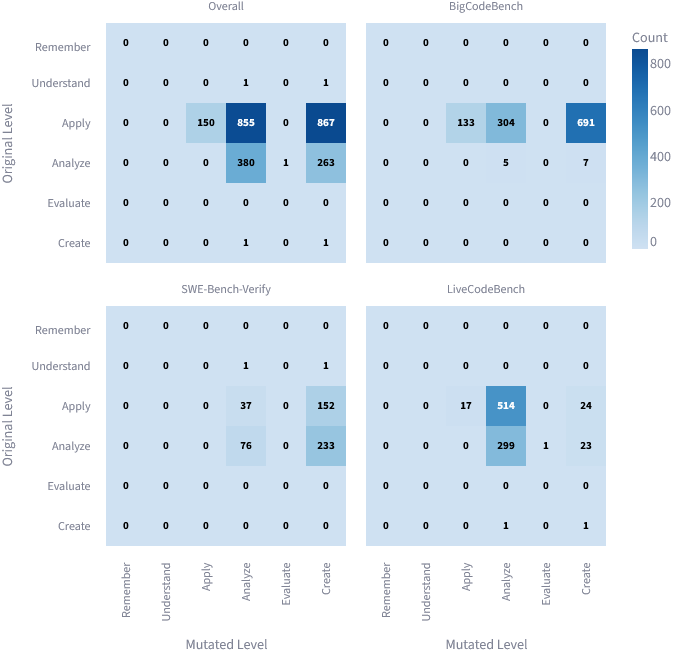}
    \\ \small Harder Intervention
\end{minipage}
\hfill
\begin{minipage}{0.48\textwidth}
    \centering
    \includegraphics[width=\textwidth]{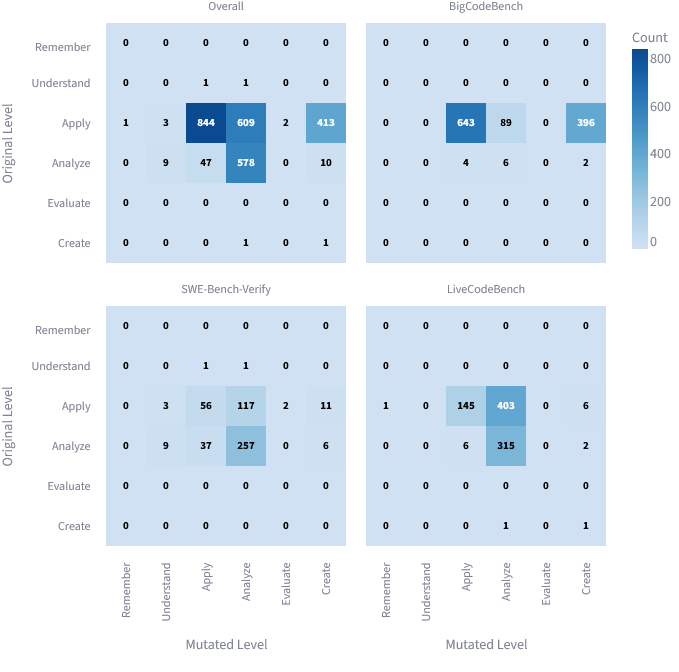}
    \\ \small Easier Intervention
\end{minipage}
\\
\vspace{0.5cm}
\begin{minipage}{0.48\textwidth}
    \centering
    \includegraphics[width=\textwidth]{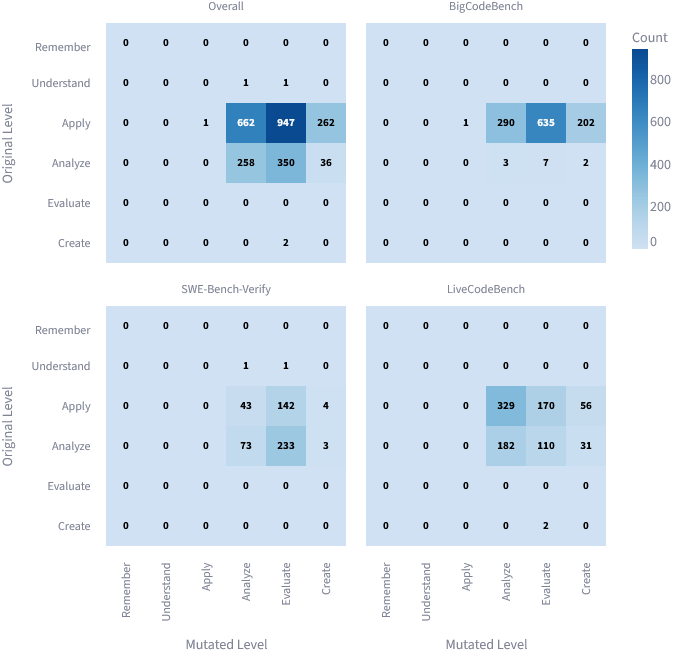}
    \\ \small Higher Target
\end{minipage}
\hfill
\begin{minipage}{0.48\textwidth}
    \centering
    \includegraphics[width=\textwidth]{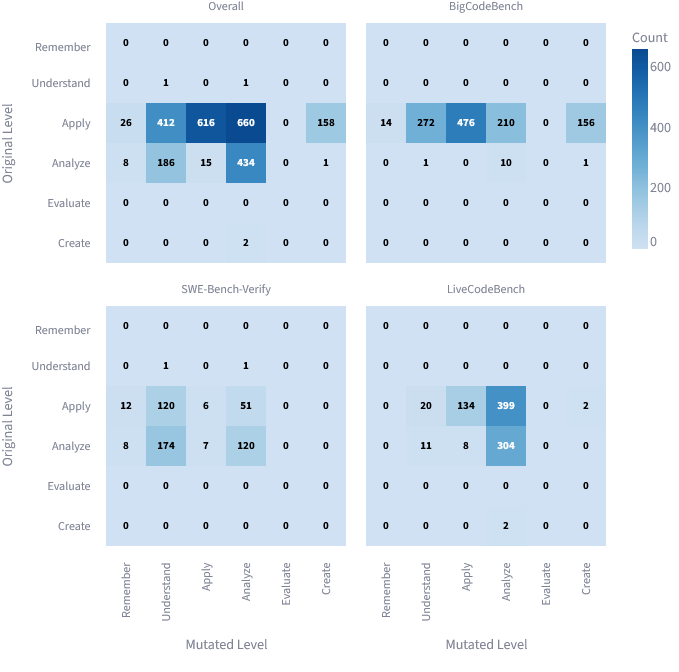}
    \\ \small Lower Target
\end{minipage}
\caption{Transition heatmaps for Qwen3-Coder-Next across different intervention settings.}
\label{fig:transition_coder}
\end{figure}

To illustrate how task mutations alter cognitive complexity, Figures \ref{fig:transition_instruct} and \ref{fig:transition_coder} map the transition dynamics of Bloom's levels for the general and coder models. The resulting heatmaps highlight distinct behavioral patterns driven by both general difficulty control and Bloom's control.

\clearpage
\section{Examples of Downward Mutations}
\label{app:downward_examples}

The examples below show representative downward mutations generated by Qwen3-Next-80B-A3B-Instruct. The labels in Table~\ref{tab:downward_examples} are the original and mutated Bloom labels for these examples.

\begin{tcolorbox}[title=Example 1: \texttt{1873\_A}, fonttitle=\bfseries, colback=gray!5!white, colframe=gray!75!black]
\small
\textbf{Original task.} There are three cards with letters \texttt{a}, \texttt{b}, \texttt{c} placed in a row in some order. You may pick two cards and swap them at most once. Determine whether the row can become \texttt{abc}; output ``YES'' if possible and ``NO'' otherwise.

\medskip
\textbf{Lower mutation.} There are three cards with letters \texttt{a}, \texttt{b}, \texttt{c} placed in a row. You can swap at most two cards once. Which of the following arrangements can become \texttt{abc} after at most one swap?
\end{tcolorbox}

\begin{tcolorbox}[title=Example 2: \texttt{astropy-14365}, fonttitle=\bfseries, colback=gray!5!white, colframe=gray!75!black]
\small
\textbf{Original task.} Please fix this issue: \texttt{ascii.qdp} Table format assumes QDP commands are upper case. The issue states that QDP commands such as \texttt{READ SERR 1 2} should also work as lowercase \texttt{read serr 1 2}, and the expected behavior is that the file reads into a \texttt{Table} with errors rather than raising \texttt{ValueError}.

\medskip
\textbf{Lower mutation.} What is the expected behavior when a QDP file uses lowercase commands like \texttt{read serr 1 2} instead of uppercase ones like \texttt{READ SERR 1 2}? Output: The table should be read successfully without raising a \texttt{ValueError}.
\end{tcolorbox}

\begin{table}[htbp]
\centering
\small
\begin{tabular}{@{}p{0.18\textwidth}p{0.22\textwidth}p{0.50\textwidth}@{}}
\toprule
ID & Labels & Interpretation \\ \midrule
\texttt{1873\_A} & Lower: Apply $\rightarrow$ Analyze & The mutation is shorter and more explicitly formatted, but it still asks whether an arrangement can become \texttt{abc} after at most one swap. The solver must analyze permutation reachability rather than recall a stated fact, so the surface format changes while the cognitive operation remains high. \\ \addlinespace
\texttt{astropy-14365} & Lower: Apply $\rightarrow$ Understand & The original SWE-Bench-Verified issue already states the expected behavior for lowercase QDP commands. The lower mutation asks for this expected behavior directly, so issue-provided information can become an Understand-level question. \\
\bottomrule
\end{tabular}
\caption{Representative examples showing how downward mutations can either preserve the required reasoning operation or use information already present in a repository issue to create a lower-level question.}
\label{tab:downward_examples}
\end{table}

\clearpage
\section{HDBSCAN-Derived Keyword Clusters}
\label{app:keywords}

The following tables report the HDBSCAN-derived keyword clusters for the semantic-delta analysis across both intervention settings and both models.

\begin{table}[htbp]
\centering
\resizebox{0.95\textwidth}{!}{%
\begin{tabular}{@{}lllll@{}}
\toprule
Mutation & Benchmark & n\_clusters & n\_noise & Keywords \\ \midrule
\multirow{4}{*}{Harder} & All & 9 & 841 & \begin{tabular}[c]{@{}l@{}}cluster2 (365): optional, bool, non, int, timeout \\ cluster0 (336): non, including, custom, nested, correctly \\ cluster1 (270): float, non, numeric, original, nan \\ cluster4 (233): valid, non, exactly, distinct, sequence \\ cluster3 (196): true, sum, valid, non, total \\ cluster5 (181): float, noise\_level, noise, non, valid \\ cluster7 (64): true, threshold, false, float, bool \\ cluster6 (19): regularization, preprocessing, float, degree, pipeline \\ cluster8 (15): weighted, 10, weights, column, entries\end{tabular} \\ \cmidrule(l){2-5} 
  & BigCodeBench & 9 & 393 & \begin{tabular}[c]{@{}l@{}}cluster2 (260): false, optional, bool, int, timeout \\ cluster3 (155): order, true, valid, non, false \\ cluster1 (110): float, non, false, outlier\_threshold, outliers \\ cluster5 (84): float, noise\_level, noise, non, gaussian \\ cluster7 (64): true, threshold, false, float, bool \\ cluster4 (22): constraints, true, valueerror, rows, contain \\ cluster6 (19): regularization, preprocessing, float, degree, pipeline \\ cluster0 (18): 2023, non, true, times, dst \\ cluster8 (15): weighted, 10, weights, column, entries\end{tabular} \\ \cmidrule(l){2-5} 
  & SWE-Bench-Verified & 3 & 78 & \begin{tabular}[c]{@{}l@{}}cluster0 (288): non, nested, including, custom, correctly \\ cluster1 (68): class, non, implement, original, dimensions \\ cluster2 (66): symbolic, expression, implement, non, correctly\end{tabular} \\ \cmidrule(l){2-5} 
  & LiveCodeBench & 6 & 370 & \begin{tabular}[c]{@{}l@{}}cluster4 (211): distinct, exactly, valid, sequence, strictly \\ cluster5 (97): valid, exactly, set, forbidden, pairs \\ cluster1 (92): exactly, distinct, pair, sum, valid \\ cluster3 (41): sum, total, weight, weighted, let \\ cluster2 (39): let, prime, divisible, xor, index \\ cluster0 (30): query, queries, vertex, lines, path\end{tabular} \\ \midrule
\multirow{4}{*}{Easier} & All & 10 & 921 & \begin{tabular}[c]{@{}l@{}}cluster1 (838): list, function, numbers, takes, count \\ cluster3 (195): numbers, function, list, takes, use \\ cluster0 (128): function, write, takes, list, number \\ cluster9 (108): write, folder, function, takes, txt \\ cluster5 (83): write, given, function, takes, age \\ cluster2 (78): write, function, given, use, expression \\ cluster4 (68): list, dataframe, 01, pd, column \\ cluster7 (57): function, string, write, takes, hello \\ cluster8 (26): hello, function, write, world, apple \\ cluster6 (18): write, takes, function, list, value\end{tabular} \\ \cmidrule(l){2-5} 
  & BigCodeBench & 10 & 515 & \begin{tabular}[c]{@{}l@{}}cluster3 (154): numbers, function, takes, list, 10 \\ cluster9 (108): write, folder, function, takes, txt \\ cluster5 (83): write, given, function, takes, age \\ cluster4 (68): list, dataframe, 01, pd, column \\ cluster7 (57): function, string, write, takes, hello \\ cluster1 (43): takes, use, axis, line, 06 \\ cluster0 (41): takes, numbers, 2023, function, write \\ cluster2 (27): use, create, product, numbers, named \\ cluster8 (26): hello, function, write, world, apple \\ cluster6 (18): write, takes, function, list, value\end{tabular} \\ \cmidrule(l){2-5} 
  & SWE-Bench-Verified & 4 & 187 & \begin{tabular}[c]{@{}l@{}}cluster1 (174): list, function, returns, numbers, write \\ cluster2 (51): function, matrix, write, expression, given \\ cluster0 (47): function, write, returns, string, hello \\ cluster3 (41): model, test, models, class, use\end{tabular} \\ \cmidrule(l){2-5} 
  & LiveCodeBench & 2 & 219 & \begin{tabular}[c]{@{}l@{}}cluster1 (621): count, exactly, number, given, list \\ cluster0 (40): number, connected, total, vertices, count\end{tabular} \\ \bottomrule
\end{tabular}%
}
\caption{HDBSCAN clustering results for general difficulty mutations generated by Qwen3-Next-80B-A3B, showing the top 5 keywords per cluster. Cluster sizes are indicated in parentheses and sorted in descending order.}
\label{tab:keyword_general_he}
\end{table}

\begin{table}[htbp]
\centering
\resizebox{\textwidth}{!}{%
\begin{tabular}{@{}lllll@{}}
\toprule
Mutation & Benchmark & n\_clusters & n\_noise & Keywords \\ \midrule
\multirow{4}{*}{Higher} & All & 8 & 546 & \begin{tabular}[c]{@{}l@{}}cluster1 (1286): based, score, evaluate, evaluates, statistical \\ cluster0 (244): time, algorithm, edge, method, based \\ cluster2 (182): solution, behavior, design, implementation, cases \\ cluster7 (86): optimal, strategy, design, multiple, yield \\ cluster3 (82): strategy, cost, optimal, solution, new \\ cluster4 (40): original, optimal, total, target, based \\ cluster5 (34): optimal, greedy, strategy, approach, valid \\ cluster6 (20): maximum, possible, edge, distinct, determine\end{tabular} \\ \cmidrule(l){2-5} 
  & BigCodeBench & 2 & 16 & \begin{tabular}[c]{@{}l@{}}cluster1 (1107): based, score, statistical, evaluates, user \\ cluster0 (17): algorithm, based, silhouette, dbscan, score\end{tabular} \\ \cmidrule(l){2-5} 
  & SWE-Bench-Verified & 4 & 148 & \begin{tabular}[c]{@{}l@{}}cluster2 (151): solution, behavior, implementation, design, performance \\ cluster1 (143): evaluate, behavior, solution, cases, design \\ cluster0 (38): solution, api, documentation, cases, type \\ cluster3 (20): test, behavior, user, unittest, solution\end{tabular} \\ \cmidrule(l){2-5} 
  & LiveCodeBench & 8 & 382 & \begin{tabular}[c]{@{}l@{}}cluster0 (189): time, edge, space, cases, terms \\ cluster7 (86): optimal, strategy, design, multiple, yield \\ cluster3 (62): strategy, optimal, cost, new, minimizes \\ cluster4 (40): original, optimal, total, target, based \\ cluster1 (36): original, sequence, set, valid, design \\ cluster5 (34): optimal, greedy, strategy, approach, valid \\ cluster2 (31): valid, sequence, exists, distinct, determine \\ cluster6 (20): maximum, possible, edge, distinct, determine\end{tabular} \\ \midrule
\multirow{4}{*}{Lower} & All & 8 & 455 & \begin{tabular}[c]{@{}l@{}}cluster1 (1119): given, write, following, identify, function \\ cluster0 (702): function, does, given, purpose, task\_func \\ cluster5 (88): sequence, position, number, given, numbered \\ cluster4 (43): following, given, valid, true, different \\ cluster3 (42): given, integers, positions, position, calculated \\ cluster7 (31): number, position, type, operations, sequence \\ cluster6 (24): abc, string, given, position, positions \\ cluster2 (16): given, integers, positive, number, representing\end{tabular} \\ \cmidrule(l){2-5} 
  & BigCodeBench & 2 & 9 & \begin{tabular}[c]{@{}l@{}}cluster1 (1008): given, write, following, identify, function \\ cluster0 (123): function, given, does, task\_func, purpose\end{tabular} \\ \cmidrule(l){2-5} 
  & SWE-Bench-Verified & 2 & 5 & \begin{tabular}[c]{@{}l@{}}cluster0 (473): does, using, method, purpose, function \\ cluster1 (22): does, string, format, request, true\end{tabular} \\ \cmidrule(l){2-5} 
  & LiveCodeBench & 8 & 441 & \begin{tabular}[c]{@{}l@{}}cluster0 (106): following, valid, true, numbers, called \\ cluster1 (89): true, following, happens, vertex, placed \\ cluster5 (88): sequence, position, number, given, numbered \\ cluster4 (43): following, given, valid, true, different \\ cluster3 (42): given, integers, positions, position, calculated \\ cluster7 (31): number, position, type, operations, sequence \\ cluster6 (24): abc, string, given, position, positions \\ cluster2 (16): given, integers, positive, number, representing\end{tabular} \\ \bottomrule
\end{tabular}%
}
\caption{HDBSCAN clustering results for Bloom's-control mutations generated by Qwen3-Next-80B-A3B, showing the top 5 keywords per cluster. Cluster sizes are indicated in parentheses and sorted in descending order.}
\label{tab:keyword_general_hl}
\end{table}

\begin{table}[htbp]
\centering
\resizebox{\textwidth}{!}{%
\begin{tabular}{@{}lllll@{}}
\toprule
Mutation & Benchmark & n\_clusters & n\_noise & Keywords \\ \midrule
\multirow{4}{*}{Harder} & All & 3 & 137 & \begin{tabular}[c]{@{}l@{}}cluster1 (1979): valid, handle, non, additionally, compute \\ cluster0 (357): implement, cases, correctly, including, handle \\ cluster2 (47): list, handle, str, string, int\end{tabular} \\ \cmidrule(l){2-5} 
  & BigCodeBench & 2 & 0 & \begin{tabular}[c]{@{}l@{}}cluster1 (986): using, handle, non, following, bool \\ cluster0 (154): false, 03, world, lowercase, hello\end{tabular} \\ \cmidrule(l){2-5} 
  & SWE-Bench-Verified & 3 & 89 & \begin{tabular}[c]{@{}l@{}}cluster1 (184): custom, cases, correctly, handle, solution \\ cluster0 (180): implement, cases, correctly, non, implementation \\ cluster2 (47): list, handle, str, string, int\end{tabular} \\ \cmidrule(l){2-5} 
  & LiveCodeBench & 2 & 48 & \begin{tabular}[c]{@{}l@{}}cluster1 (809): valid, exactly, additionally, define, let \\ cluster0 (23): valid, super, total, digits, false\end{tabular} \\ \midrule
\multirow{4}{*}{Easier} & All & 6 & 773 & \begin{tabular}[c]{@{}l@{}}cluster1 (597): function, takes, given, showing, use \\ cluster2 (549): contains, function, integers, write, constraints \\ cluster0 (387): given, list, function, number, integers \\ cluster3 (76): function, write, use, representing, takes \\ cluster4 (76): function, write, given, takes, integer \\ cluster5 (62): function, write, tmp, given, use\end{tabular} \\ \cmidrule(l){2-5} 
  & BigCodeBench & 6 & 336 & \begin{tabular}[c]{@{}l@{}}cluster1 (511): function, takes, given, use, showing \\ cluster3 (76): function, write, use, representing, takes \\ cluster4 (76): function, write, given, takes, integer \\ cluster5 (62): function, write, tmp, given, use \\ cluster2 (44): function, times, write, takes, value \\ cluster0 (35): function, takes, given, use, write\end{tabular} \\ \cmidrule(l){2-5} 
  & SWE-Bench-Verified & 3 & 88 & \begin{tabular}[c]{@{}l@{}}cluster2 (327): model, function, field, correctly, using \\ cluster1 (64): outerref, filter, exclude, results, qs \\ cluster0 (21): function, using, documentation, signature, error\end{tabular} \\ \cmidrule(l){2-5} 
  & LiveCodeBench & 3 & 349 & \begin{tabular}[c]{@{}l@{}}cluster0 (331): given, list, number, exactly, line \\ cluster2 (178): contains, line, integers, 100, constraints \\ cluster1 (22): lines, contains, need, grid, line\end{tabular} \\ \bottomrule
\end{tabular}%
}
\caption{HDBSCAN clustering results for general difficulty mutations generated by Qwen3-Coder-Next, showing the top 5 keywords per cluster. Cluster sizes are indicated in parentheses and sorted in descending order.}
\label{tab:keyword_coder_he}
\end{table}

\begin{table}[htbp]
\centering
\resizebox{\textwidth}{!}{%
\begin{tabular}{@{}lllll@{}}
\toprule
Mutation & Benchmark & n\_clusters & n\_noise & Keywords \\ \midrule
\multirow{4}{*}{Higher} & All & 5 & 945 & \begin{tabular}[c]{@{}l@{}}cluster1 (572): uses, implementation, evaluate, given, based \\ cluster0 (425): implementation, uses, behavior, based, given \\ cluster2 (257): strategy, optimal, function, valid, determine \\ cluster3 (179): function, edge, evaluate, implementation, cases \\ cluster4 (141): time, space, algorithm, complexity, evaluate\end{tabular} \\ \cmidrule(l){2-5} 
  & BigCodeBench & 4 & 369 & \begin{tabular}[c]{@{}l@{}}cluster1 (317): uses, implementation, given, based, python \\ cluster0 (216): uses, implementation, given, based, implementations \\ cluster3 (159): function, evaluate, implementation, cases, design \\ cluster2 (79): function, following, implementation, cases, specific\end{tabular} \\ \cmidrule(l){2-5} 
  & SWE-Bench-Verified & 2 & 92 & \begin{tabular}[c]{@{}l@{}}cluster1 (217): behavior, evaluate, current, dtype, da \\ cluster0 (191): behavior, current, implementation, evaluation, evaluate\end{tabular} \\ \cmidrule(l){2-5} 
  & LiveCodeBench & 5 & 484 & \begin{tabular}[c]{@{}l@{}}cluster2 (178): strategy, optimal, valid, determine, given \\ cluster4 (141): time, space, algorithm, complexity, evaluate \\ cluster1 (38): true, false, valid, claim, determine \\ cluster3 (20): solution, dp, uses, algorithm, justify \\ cluster0 (18): approach, time, sequence, solution, complexity\end{tabular} \\ \midrule
\multirow{4}{*}{Lower} & All & 5 & 413 & \begin{tabular}[c]{@{}l@{}}cluster2 (1104): given, function, write, takes, use \\ cluster1 (653): given, does, function, purpose, method \\ cluster0 (199): function, code, provided, purpose, does \\ cluster4 (125): given, list, grid, position, sequence \\ cluster3 (25): s\_, contains, line, constraints, integers\end{tabular} \\ \cmidrule(l){2-5} 
  & BigCodeBench & 3 & 58 & \begin{tabular}[c]{@{}l@{}}cluster2 (900): given, function, write, following, resulting \\ cluster0 (164): function, code, provided, does, task\_func \\ cluster1 (18): function, python, required, libraries, modules\end{tabular} \\ \cmidrule(l){2-5} 
  & SWE-Bench-Verified & 2 & 4 & \begin{tabular}[c]{@{}l@{}}cluster1 (478): method, purpose, function, using, happens \\ cluster0 (18): documentation, using, autodoc\_typehints, description, section\end{tabular} \\ \cmidrule(l){2-5} 
  & LiveCodeBench & 5 & 351 & \begin{tabular}[c]{@{}l@{}}cluster2 (204): given, list, exactly, positive, need \\ cluster1 (157): given, does, goal, definition, integer \\ cluster4 (125): given, list, grid, position, sequence \\ cluster3 (25): s\_, contains, line, constraints, integers \\ cluster0 (17): contain, given, length, sequence, element\end{tabular} \\ \bottomrule
\end{tabular}%
}
\caption{HDBSCAN clustering results for Bloom's-control mutations generated by Qwen3-Coder-Next, showing the top 5 keywords per cluster. Cluster sizes are indicated in parentheses and sorted in descending order.}
\label{tab:keyword_coder_hl}
\end{table}

\clearpage
\section{PCA of Layer Embeddings}
\label{app:pca_embeddings}

The PCA plots below provide a compact view of the latent geometry underlying the behavioral results. They project layer embeddings from the depths where each model reached peak linear separability, making it easier to compare how the general model and the coder model encode general difficulty control and Bloom's control.

\begin{figure}[ht!]
  \centering
  \includegraphics[width=0.7\textwidth]{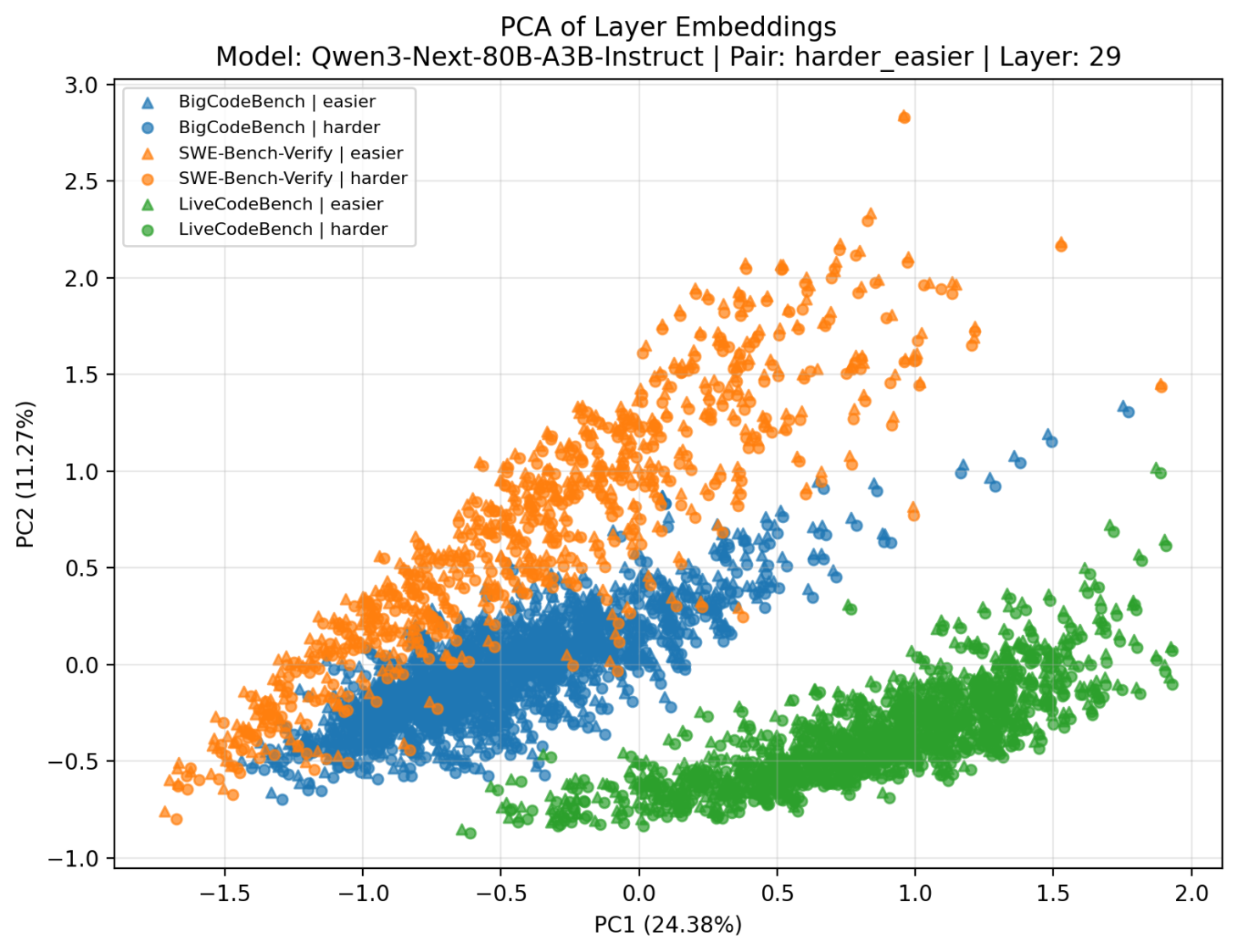}
  \caption{PCA projection of layer embeddings for the general model under general difficulty control (Harder versus Easier).}
  \label{fig:pca_general_harder_easier}
\end{figure}

\begin{figure}[ht!]
  \centering
  \includegraphics[width=0.7\textwidth]{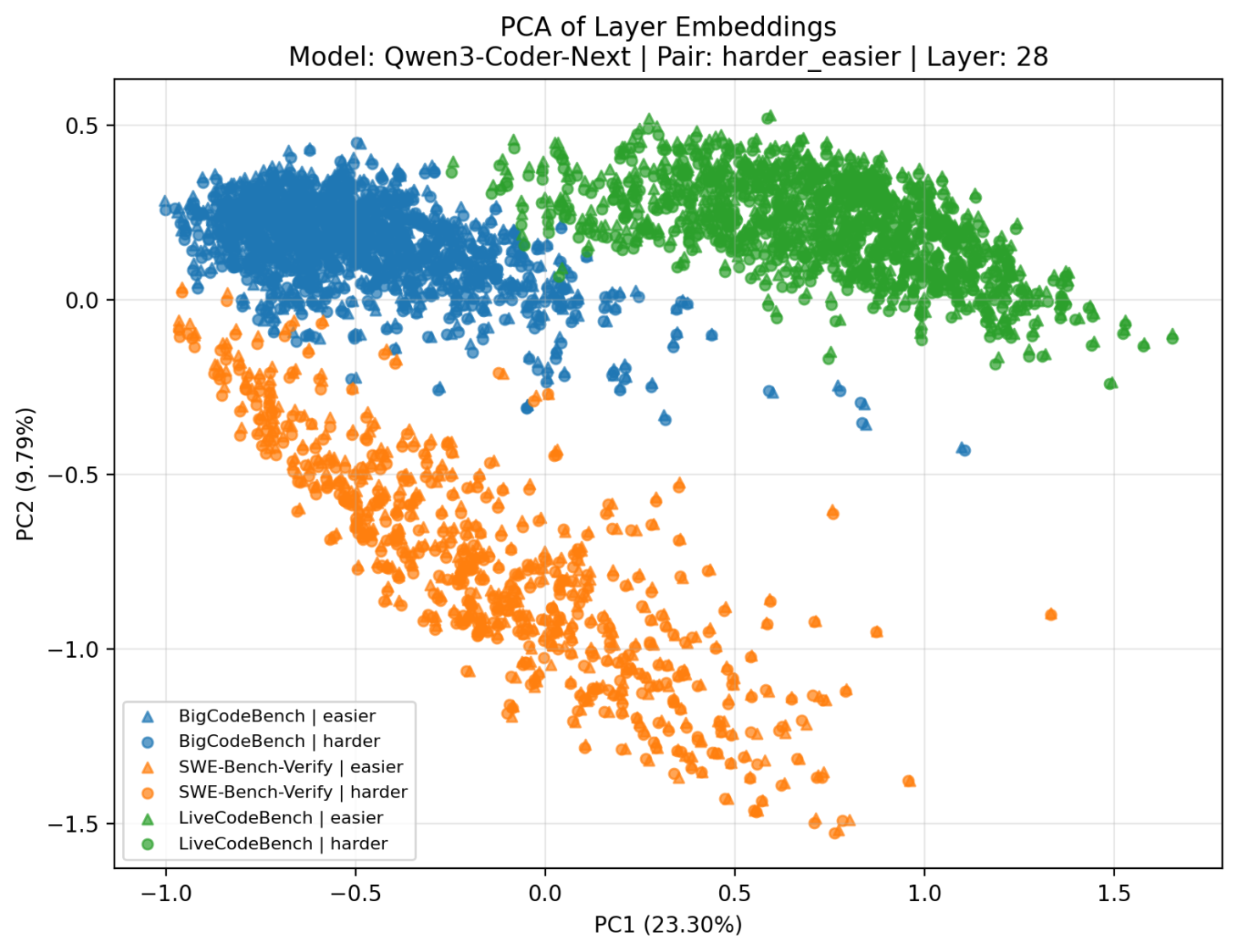}
  \caption{PCA projection of layer embeddings for the coder model under general difficulty control (Harder versus Easier).}
  \label{fig:pca_coder_harder_easier}
\end{figure}

\begin{figure}[ht!]
  \centering
  \includegraphics[width=0.7\textwidth]{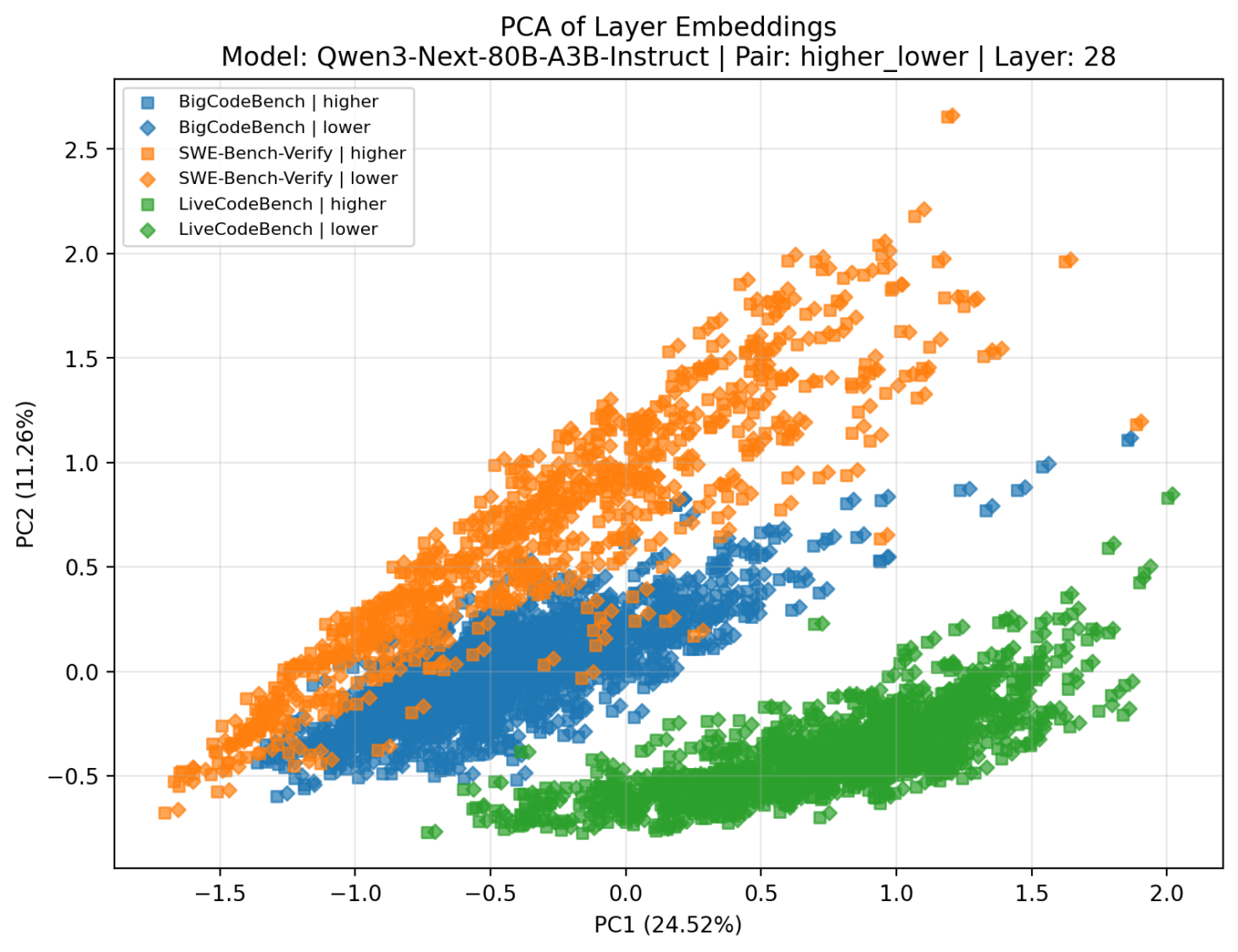}
  \caption{PCA projection of layer embeddings for the general model under Bloom's control (Higher versus Lower).}
  \label{fig:pca_general_higher_lower}
\end{figure}

\begin{figure}[ht!]
  \centering
  \includegraphics[width=0.7\textwidth]{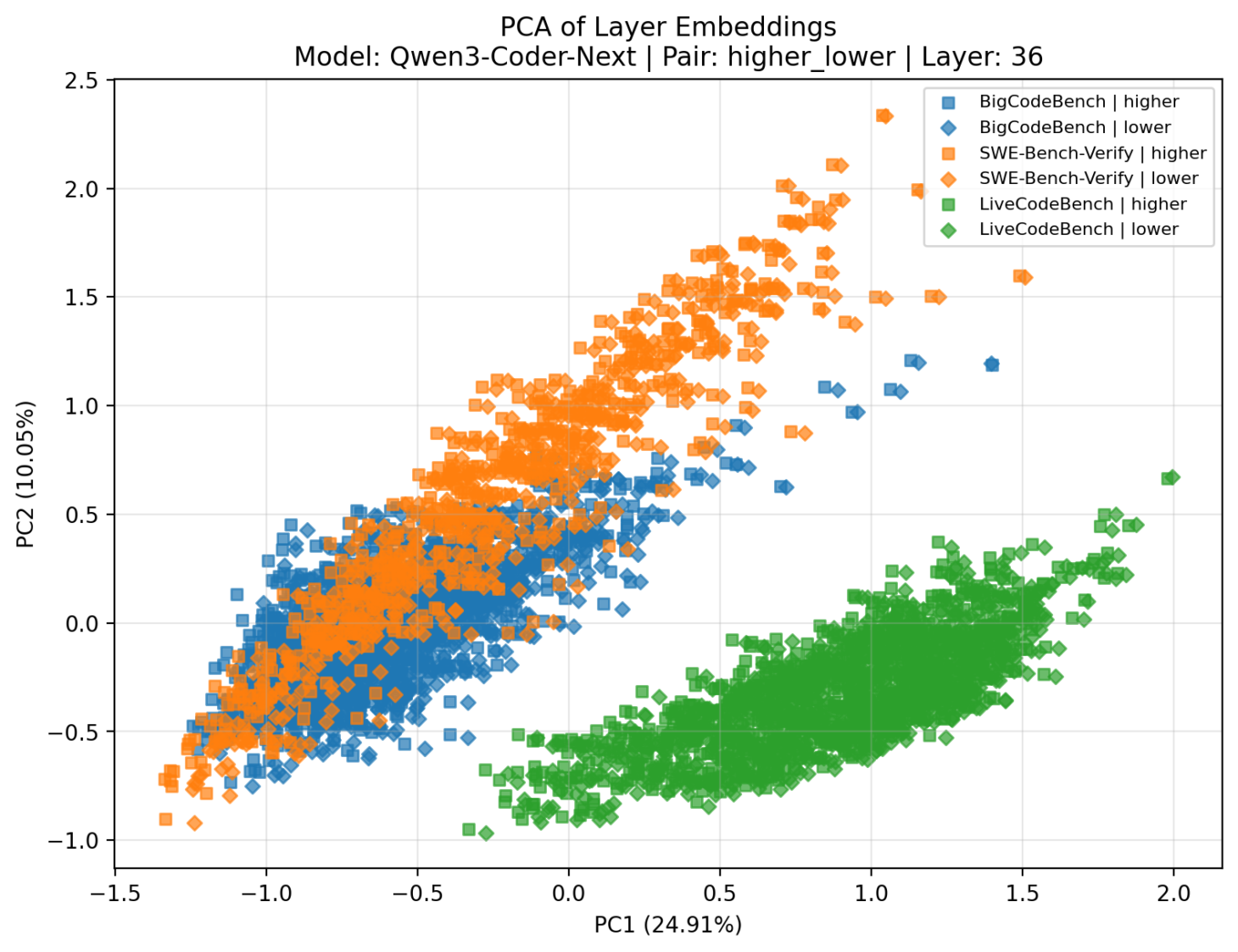}
  \caption{PCA projection of layer embeddings for the coder model under Bloom's control (Higher versus Lower).}
  \label{fig:pca_coder_higher_lower}
\end{figure}

\end{document}